\documentclass[10pt,twocolumn,letterpaper]{article}

\PassOptionsToPackage{table}{xcolor}
\usepackage[pagenumbers]{cvpr} 

\usepackage{array,multirow}
\usepackage{float}
\usepackage{graphicx}
\usepackage{subcaption}
\usepackage{placeins}

\definecolor{lightgray}{gray}{0.95}
\definecolor{gray}{gray}{0.8}
\colorlet{lightblue}{blue!20}
\colorlet{lightgreen}{OliveGreen!20}
\colorlet{lightred}{red!20}
\colorlet{darkblue}{blue!30}
\colorlet{darkgreen}{OliveGreen!30}
\colorlet{darkred}{red!30}

\definecolor{cvprblue}{rgb}{0.21,0.49,0.74}
\usepackage[pagebackref,breaklinks,colorlinks,allcolors=cvprblue]{hyperref}

\def\paperID{17} 
\def\confName{CVPR}
\def\confYear{2025}

\title{Human Aligned Compression for Robust Models}

\author{Samuel Räber\\
ETH Zürich\\
{\tt\small sraeber@student.ethz.ch}
\and
Andreas Plesner\\
ETH Zürich\\
{\tt\small aplesner@ethz.ch}
\and
Till Aczel\\
ETH Zürich\\
{\tt\small taczel@ethz.ch}
\and
Roger Wattenhofer\\
ETH Zürich\\
{\tt\small wattenhofer@ethz.ch}
}

\begin{document}
\maketitle
\begin{abstract}
    Adversarial attacks on image models threaten system robustness by introducing imperceptible perturbations that cause incorrect predictions. 
    We investigate human-aligned learned lossy compression as a defense mechanism, comparing two learned models (HiFiC and ELIC) against traditional JPEG across various quality levels. 
    Our experiments on ImageNet subsets demonstrate that learned compression methods outperform JPEG, particularly for Vision Transformer architectures, by preserving semantically meaningful content while removing adversarial noise. 
    Even in white-box settings where attackers can access the defense, these methods maintain substantial effectiveness. 
    We also show that sequential compression—applying rounds of compression/decompression—significantly enhances defense efficacy while maintaining classification performance. 
    Our findings reveal that human-aligned compression provides an effective, computationally efficient defense that protects the image features most relevant to human and machine understanding. It offers a practical approach to improving model robustness against adversarial threats.\footnote{\url{https://github.com/aplesner/Human-aligned-compression-for-robust-models}.}
\end{abstract}

\section{Introduction}
\label{sec:intro}

Vision models have made significant improvements in recent years, achieving remarkable success in tasks like image classification \cite{he2015deepresiduallearningimage}, object detection \cite{NIPS2013_f7cade80,10028728} and medical imaging \cite{shenmedical2017}. 
However, vision models remain highly vulnerable to adversarial attacks despite these advancements. 
Adversarial attacks are carefully crafted perturbations added to input images, which are often imperceptible to the human eye but can cause deep learning models to make incorrect predictions \cite{szegedy2014intriguingpropertiesneuralnetworks,madry2019deeplearningmodelsresistant}.
These attacks pose a serious threat to applications that rely on the reliability of vision models, such as autonomous driving, healthcare diagnostics, and surveillance systems.

One potential approach to defending against these adversarial attacks is to remove the small perturbations introduced by the attacker. 
Eliminating these imperceptible changes makes it possible to prevent the model from being misled. 
Traditional methods, such as blurring or adding random noise, can be effective at removing perturbations; however, they have significant drawbacks. 
While these methods reduce the adversarial impact, they distort the image, changing its distribution in a way that may cause the image to be ``out of distribution" for the classifier. 
Moreover, they do not solely remove adversarial perturbations—they also remove information that could be important for the task. 
This results in a loss of critical information, degrading model performance and making the system less reliable.

\citet{ilyas2019adversarial} showed in their seminal work that image classifiers learn to use non-robust features for image classification that the adversarial examples exploit. In the area of image compression, researchers showed that human-perception-aligned learned lossy compression models can yield high compression ratios while producing images that humans prefer, for instance, over JPEG compressed images. 
These techniques aim to preserve an image's most important features according to human perception while discarding less significant details. 
By doing so, they can remove adversarial perturbations without altering the underlying image distribution and without losing task-relevant information. 
Since the images remain in distribution for the classifier, the model can continue to perform effectively while being protected from adversarial attacks. \citet{dziugaite2016studyeffectjpgcompression} has shown that JPEG compression could be a viable defence, however, as later shown, only if the attacker does not include the JPEG compression in the attack \cite{shin2017jpeg}. 

This work explores the potential of learned image compression methods to defend against adversarial attacks. 
We compare these methods to a traditional technique, JPEG, and evaluate their effectiveness in removing adversarial perturbations while preserving the integrity of the image distribution. 
Our findings show that human-perception aligned compression offers a promising strategy for defending vision models against adversarial attacks, without sacrificing classification accuracy. 
This approach contributes to developing more efficient and robust defense mechanisms, fostering the creation of more secure and reliable vision systems.

\section{Related Work}
\label{sec:related_work}
\subsection{Adversarial attacks}

Adversarial examples have become a critical concern in machine learning, particularly regarding the robustness and security of deep learning models.
These inputs are intentionally crafted to deceive models into making incorrect predictions. \citet{szegedy2014intriguingpropertiesneuralnetworks} first demonstrated that small, imperceptible perturbations could cause neural networks to misclassify images with high confidence.

\citet{kurakin2017adversarialexamplesphysicalworld} extended the study of adversarial attacks to the physical world, showing that printed images with adversarial perturbations could still deceive classifiers when captured through a camera. 
This work underscored the real-world implications of adversarial attacks beyond purely digital environments.

In response to these challenges, various defense mechanisms have been proposed. 
\citet{madry2019deeplearningmodelsresistant} introduced Projected Gradient Decent (PGD) attacks along with adversarial training, a technique in which models are trained on adversarial examples to improve their robustness.
\citet{papernot2016distillationdefenseadversarialperturbations} proposed defensive distillation, leveraging knowledge distillation \cite{hinton2015distillingknowledgeneuralnetwork}—a technique that compresses an ensemble of models into a smaller model—to enhance resistance against adversarial attacks.
Another approach involves preprocessing inputs to remove adversarial perturbations before classification, which can be achieved using image compression techniques \cite{dziugaite2016studyeffectjpgcompression}. 
Despite these efforts, achieving a comprehensive defense against adversarial attacks remains an open problem. 

\subsection{Image compression as adversarial defense}

\citet{dziugaite2016studyeffectjpgcompression} demonstrated that JPEG compression can weaken adversarial attacks by removing small perturbations.
The perturbations often vanish by compressing and decompressing a potentially manipulated image, reducing the attack's effectiveness.
Even with a high quality factor of 75, JPEG compression enhanced model robustness.

However, \citet{shin2017jpeg} demonstrated that the defensive effectiveness of JPEG compression can be significantly diminished by leveraging a differentiable approximation of the algorithm. 
By propagating gradient information through the model \emph{and} the JPEG compression process, adversarial attacks can generate perturbations that persist even after the compression and decompression steps.

\subsection{Learned lossy image compression}

Recent advancements in image compression have moved beyond traditional methods like JPEG, which use linear transformations, to learned techniques that replace the Discrete Cosine Transform (DCT) with nonlinear transformations \cite{LiU_2023_CVPR,He_2021_CVPR,galteri2019deep,agustsson2019generativeadversarialnetworksextreme,minnen2018jointautoregressivehierarchicalpriors}.
Variational Autoencoder (VAE)-based models and Generative Adversarial Networks (GANs) \cite{mentzer2020highfidelitygenerativeimagecompression} help minimize compression artifacts, producing realistic images even at ultra-low bitrates. 

HiFiC \cite{mentzer2020highfidelitygenerativeimagecompression} leverages GANs to achieve visually appealing reconstructions while preserving perceptually significant information. A user study demonstrated HiFiC’s superiority in reconstruction quality over other methods, even at half the bits per pixel.

ELIC \cite{he2022elicefficientlearnedimage} optimizes image compression for both speed and efficiency. It outperforms previous learned methods, such as Minnen \etal \cite{minnen2018jointautoregressivehierarchicalpriors} and Cheng \etal \cite{cheng2020learnedimagecompressiondiscretized}, particularly at low bitrates.


At low bitrates, learned compression methods outperform JPEG by preserving perceptual quality and reducing visual artifacts, all while maintaining the original image distribution. Thus, using learned compression as preprocessing for neural network classifiers can enhance robustness by removing small perturbations while preserving the image distribution.

\section{Methods}
\label{sec:methods}

Our experiments' objective was to evaluate the effectiveness of image compression as a defense mechanism against adversarial attacks. We also investigated the impact of varying compression quality levels and the effects of applying multiple compression steps sequentially.

We implemented our defenses as an additional preprocessing step applied to the (perturbed) image before it was fed into the classifier. Since lossy image compression inherently removes specific details, it is expected to eliminate some of the adversarial perturbations, thereby reducing the effectiveness of the attack. We then generated adversarial perturbations for the dataset and assessed the classifier’s accuracy on the perturbed images, comparing it to the baseline accuracy before the attack.

\subsection{Defenses}

The compression methods used as defenses were JPEG, HiFiC, and ELIC. JPEG was selected because it is one of the most widely used compression algorithms and has been previously explored as a defense mechanism against adversarial attacks \cite{dziugaite2016studyeffectjpgcompression, shin2017jpeg}. Thus, JPEG serves as a baseline for comparison with the other compression techniques. To implement JPEG compression and decompression, we used the differentiable approximation provided by Kornia \cite{eriba2019kornia}.

HiFiC and ELIC are learned compression methods that utilize different architectures to achieve high image quality at low bitrates (\Cref{tab:bpp}, \cite{mentzer2020highfidelitygenerativeimagecompression,he2022elicefficientlearnedimage}). PyTorch implementations and checkpoints with pretrained weights are publicly available for both methods. \footnote{HiFiC: \url{https://github.com/Justin-Tan/high-fidelity-generative-compression}.} \footnote{ELIC: \url{https://github.com/VincentChandelier/ELiC-ReImplemetation}.}

For HiFiC and ELIC, we employed differentiable forward functions. The use of differentiable defenses allowed gradient information to propagate through the entire pipeline (including both the model and the defense mechanism). This is expected to reduce the effectiveness of the defense, as it enables the adversarial attack to adapt its perturbations to persist through the compression process. To account for this, we conducted an additional set of experiments using this stronger adaptive attack. In tables throughout this paper, this is annotated with through being true.



\subsection{Adversarial attacks}

We use these methods to compute adversarial examples:
\begin{itemize}
    \item Fast gradient sign method (FGSM) \cite{goodfellow2014explaining}.
    \item Iterative FGSM (iFGSM) \cite{kurakin2017adversarialexamplesphysicalworld}.
    \item Projected gradient descent (PGD) \cite{madry2019deeplearningmodelsresistant}.
    \item Carlini-Wagner attack (CW) \cite{carlini2017evaluatingrobustnessneuralnetworks}.
    \item DeepFool attack (DeepFool) \cite{moosavidezfooli2016deepfoolsimpleaccuratemethod}.
\end{itemize}
For FGSM, iFGSM and PGD we use different $l_\infty$ norm values (epsilon) to modulate the attack strength, for CW and DeepFool we computed the accuracy for perturbations below a specific $l_2$ norm value. For iFGSM and PGD we used 10 iterations. For all attacks, the torchattacks \cite{kim2020torchattacks} implementation was used. For a full list of hyperparameters see \Cref{tab:hyperparams} in the Appendix. When we pass the gradients through the compression model as in \cite{shin2017jpeg}, we call it as a ``white-box'' while if we do not, we call it a ``black-box'' attack.

\subsection{Models and datasets}
The experiments used two different base models: ResNet50 \cite{he2015deepresiduallearningimage}, and a Vision Transformer (ViT), specifically ViT-B/16 \cite{dosovitskiy2021imageworth16x16words}. Both models were sourced from PyTorch \cite{paszke2019pytorchimperativestylehighperformance} and initialized with pretrained ImageNet weights.

For our experiments, we used the validation split of Imagenette, a subset of ImageNet \cite{deng2009imagenet} containing 10 easily classified classes. In later experiments, the full ImageNet test split was utilized.

\subsection{Compression strength}

To determine the appropriate compression quality for our experiments, we conducted additional tests for each compression method, comparing different levels to identify an optimal quality setting. Beyond defensive strength, our choice of compression was also influenced by several factors: the impact of the defense on accuracy in the absence of an attack, comparability between different compression methods and related work, and the availability of pretrained weights for learned compression models. Training these models from scratch was beyond the scope of this study.

The parameters influencing compression strength for the different methods were as follows:
\begin{itemize}
    \item \textbf{JPEG:} The quality parameter $q\in[0,100]$ and controls the quantization strength of the algorithm, with lower values corresponding to greater compression. We compared values $q \in \{5.0,10.0,15.0,25.0,35.0,50.0,75.0,95.0\}$. Typically, values greater than 70 are considered high quality, while values below 30 result in low-quality images that may appear pixelated and blurry.
    \item \textbf{HiFiC:} Three different sets of pretrained weights were available for HiFiC: $\textbf{HiFiC}^{\textbf{low}}$, $\textbf{HiFiC}^{\textbf{med}}$ and $\textbf{HiFiC}^{\textbf{high}}$, which were trained to achieve target bitrates per pixel (BPP) of 0.14, 0.3, and 0.45, respectively.
    \item\textbf{ELIC:} Six different checkpoints were available for ELIC: [0004, 0008, 0016, 0032, 0150, 0450]. These correspond to different values of $\lambda$, the rate-controlling parameter, determining the trade-off between estimated bitrate and image reconstruction distortion (see \cite{he2022elicefficientlearnedimage} for details).
\end{itemize}

Additionally, we computed BPP values for images from the ImageNet dataset resized to $224\times224$ pixels across different compression methods, cf. \Cref{tab:bpp}. This allowed a direct comparison of size reduction between techniques.

\begin{table}[t]
    \centering
    \small
    \caption{
    Bits per pixel (BPP) measurements for different compression methods and quality settings, computed on 100 random 224×224 images from ImageNet. This table enables direct comparison of compression efficiency across JPEG, ELIC, and HiFiC methods at various quality levels.
    }
    \begin{tabular}{|c|c|}
    \hline
         \multicolumn{2}{|c|}{JPEG} \\
    \hline
    
    Quality & BPP \\
    
    \hline
         $q = 5.0$& 0.35\\         \hline
         $q = 10.0$& 0.48\\         \hline
         $q = 15.0$& 0.59\\         \hline
         $q = 25.0$& 0.78\\         \hline
         $q = 35.0$& 0.94\\         \hline
         $q = 50.0$& 1.14\\         \hline
         $q = 75.0$& 1.65\\         \hline
         $q = 95.0$& 3.80\\         \hline
    \end{tabular}
    \begin{tabular}{|c|c|}
    \hline
         \multicolumn{2}{|c|}{ELIC} \\
    \hline
    Weights & BPP\\         \hline
         $0004$& 0.06\\         \hline
         $0008$& 0.09\\         \hline
         $0016$& 0.14\\         \hline
         $0032$& 0.19\\         \hline
         $0150$& 0.42\\         \hline
         $0450$& 0.69\\         \hline
         \multicolumn{2}{c}{}\\
         \multicolumn{2}{c}{}

    \end{tabular}
    \begin{tabular}{|c|c|}
    \hline
         \multicolumn{2}{|c|}{HiFiC} \\
    \hline
    Weights & BPP\\         \hline
         low& 0.15\\         \hline
         med& 0.43\\         \hline
         high& 0.46\\         \hline
         \multicolumn{2}{c}{}\\
         \multicolumn{2}{c}{}\\
         \multicolumn{2}{c}{}\\
         \multicolumn{2}{c}{}\\
         \multicolumn{2}{c}{}
         
    \end{tabular}
    \label{tab:bpp}
\end{table}

\subsection{Sequential compression}
We also conducted experiments on the effectiveness of compressing and decompressing an image multiple times in sequence as a defense. We always propagated the gradients through the defense for these experiments to achieve a stronger attack. The experiments were conducted on Imagenette and ImageNet. For ImageNet, we used 1000 randomly sampled images to reduce computation time.

\begin{figure*}[!t]
    \centering
    \begin{subfigure}[t]{0.16\linewidth}
        \centering
        \includegraphics[width=\linewidth]{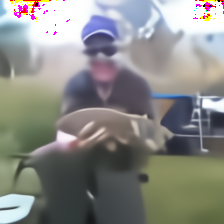}
        \caption{ELIC low quality}
        \label{fig:elic_low}
    \end{subfigure}
    \hfill
    \begin{subfigure}[t]{0.16\linewidth}
        \centering
        \includegraphics[width=\linewidth]{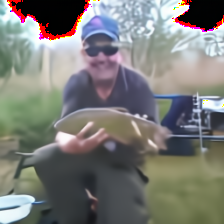}
        \caption{ELIC high quality}
        \label{fig:elic_high}
    \end{subfigure}
    \hfill
    \begin{subfigure}[t]{0.16\linewidth}
        \centering
        \includegraphics[width=\linewidth]{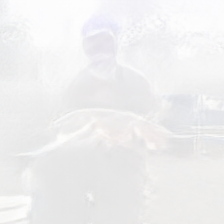}
        \caption{HiFiC low quality}
        \label{fig:hific_low}
    \end{subfigure}
    \hfill
    \begin{subfigure}[t]{0.16\linewidth}
        \centering
        \includegraphics[width=\linewidth]{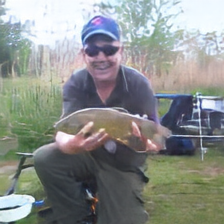}
        \caption{HiFiC med. quality}
        \label{fig:hific_med}
    \end{subfigure}
    \hfill
    \begin{subfigure}[t]{0.16\linewidth}
        \centering
        \includegraphics[width=\linewidth]{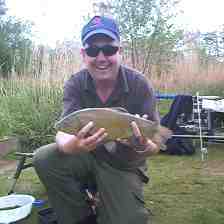}
        \caption{JPEG quality 25.0}
        \label{fig:jpeg_25}
    \end{subfigure}
    \hfill
    \begin{subfigure}[t]{0.16\linewidth}
        \centering
        \includegraphics[width=\linewidth]{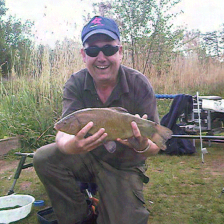}
        \caption{Original image}
        \label{fig:original}
    \end{subfigure}
    \caption{Visual comparison of image degradation after three compression/decompression cycles using different compression methods and quality settings. From left to right: (a) ELIC 0004, (b) ELIC quality 0016, (c) HiFiC low, (d) HiFiC medium, (e) JPEG quality 25.0, and (f) the original uncompressed image. Note how learned compression methods (ELIC, HiFiC) exhibit different artifact patterns than traditional JPEG compression.}
    \label{fig:images}
    \vspace{-5mm}
\end{figure*}

\section{Results}
\label{sec:results}
The main results can be found in \Cref{fig:l_inf,fig:l_2} with \Cref{tab:linf_attacks_full,tab:l2_attacks_full} in the Appendix giving the exact value. We use quality levels 25.0, low, and 0016 for the compression defenses for JPEG, HiFiC, and ELIC, respectively, unless otherwise stated.

\subsection{Baseline results}
  Both of the used models achieved a very high accuracy on Imagenette, $\approx 0.998$ for ResNet50 and $\approx 0.999$ for the ViT, as well as a strong resilience against the FGSM attack even without a defense, with both models still achieving accuracies $>0.8$ at an attack strength of epsilon $\frac{8}{255}$. For iFGSM (cf. \Cref{tab:linf_small}) and PGD the accuracy dropped to $<0.05$ for ResNet50 and $<0.01$ for the ViT at this level of attack strength. These baseline experiments also indicate a difference in robustness between the two models used, as the accuracy of the ViT is lower for all baseline experiments.
 
 \subsection{Defense results}
 \begin{figure}
     \centering
     \includegraphics[width=1.0\linewidth]{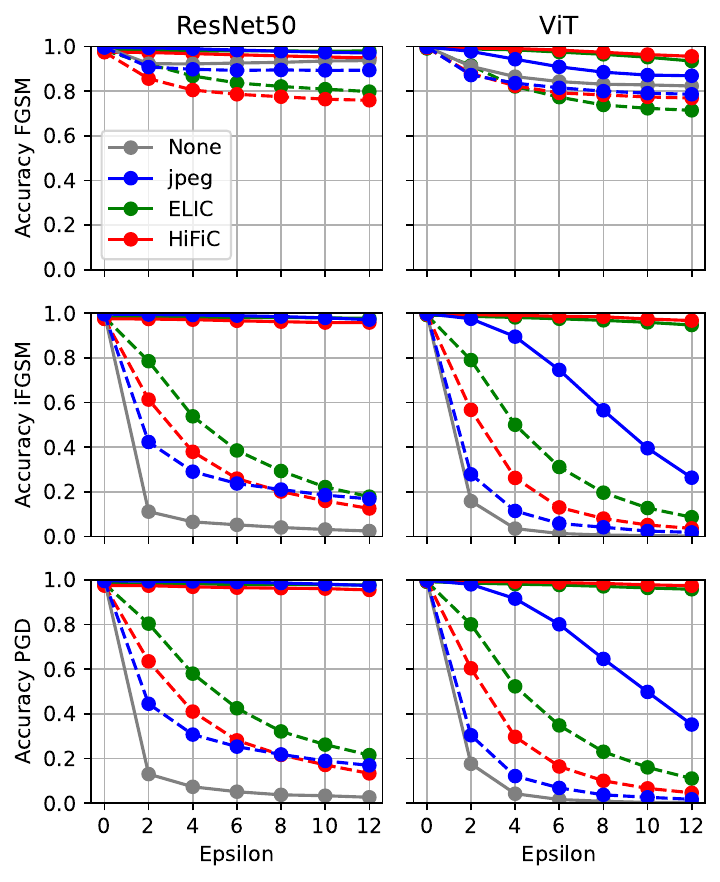}
     \caption{Model accuracy under FGSM (top row), iFGSM (middle row), and PGD (bottom row) adversarial attacks on the Imagenette dataset for ResNet50 (left) and ViT (right) architectures. Solid lines represent ``black-box'' attacks (without gradient propagation through the compression), while dashed lines show ``white-box'' attacks (with gradient propagation through the defense). Learned compression methods (ELIC, HiFiC) consistently outperform JPEG for the ViT model, particularly under stronger attacks. Epsilon values represent attack strength as $x/255$.
     }
     \label{fig:l_inf}
 \end{figure}
 \begin{figure}
     \centering
     \includegraphics[width=1.0\linewidth]{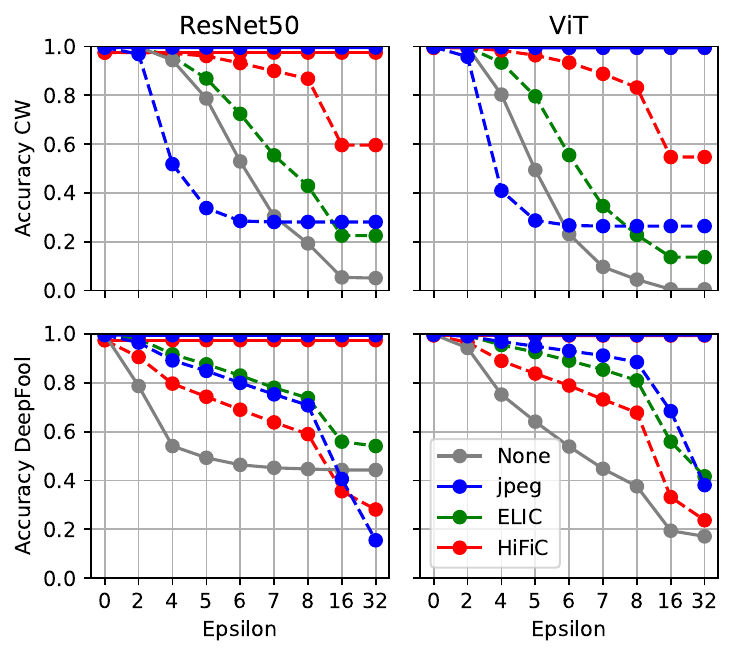}
     \caption{
     Model accuracy under Carlini-Wagner (CW, top row) and DeepFool (bottom row) adversarial attacks on the Imagenette dataset for ResNet50 (left) and ViT (right) architectures. Solid lines represent ``black-box'' attacks, while dashed lines show ``white-box'' attacks. Epsilon shows the maximum L2 norm of the perturbations. HiFiC demonstrates superior robustness against CW attacks, especially for the ViT model.
     }
     \label{fig:l_2}
 \end{figure}

\begin{table}[t]
    \centering
    \small
    \caption{
    Classification accuracy (\%) for different defenses against iFGSM attacks on the Imagenette dataset at varying epsilon values. Results are shown for both ResNet50 and ViT models under black-box (Through=False) and white-box (Through=True) attack scenarios, demonstrating the vulnerability of all defenses to white-box attacks.
    }
    \begin{tabular}{|c|c|c|c|c|c|c|}
        \toprule
         &Defense& Through & 0 & $\frac{4}{255}$ & $\frac{8}{255}$ & $\frac{12}{255}$  \\ \midrule
         \parbox[t]{2mm}{\multirow{7}{*}{\rotatebox[origin=c]{90}{ResNet50}}}  &  None  &  \cellcolor{lightgray}False &  \cellcolor{lightgray}100 &  \cellcolor{lightgray}6 &  \cellcolor{lightgray}4 &  \cellcolor{lightgray} 2 \\  \cline{2-7}
   &  \multirow{2}{*}{JPEG}  &  \cellcolor{lightblue}False &  \cellcolor{lightblue}100 &  \cellcolor{lightblue}99 &  \cellcolor{lightblue}98 &  \cellcolor{lightblue} 97 \\  \cline{3-7}
   &     &  \cellcolor{darkblue}True &  \cellcolor{darkblue}100 &  \cellcolor{darkblue}28 &  \cellcolor{darkblue}21 &  \cellcolor{darkblue} 17 \\  \cline{2-7}
   &  \multirow{2}{*}{ELIC}  &  \cellcolor{lightred}False &  \cellcolor{lightred}98 &  \cellcolor{lightred}98 &  \cellcolor{lightred}98 &  \cellcolor{lightred} 98 \\  \cline{3-7}
   &     &  \cellcolor{darkred}True &  \cellcolor{darkred}98 &  \cellcolor{darkred}54 &  \cellcolor{darkred}28 &  \cellcolor{darkred} 18 \\  \cline{2-7}
   &  \multirow{2}{*}{HiFiC}  &  \cellcolor{lightgreen}False &  \cellcolor{lightgreen}98 &  \cellcolor{lightgreen}97 &  \cellcolor{lightgreen}96 &  \cellcolor{lightgreen} 96 \\  \cline{3-7}
   &     &  \cellcolor{darkgreen}True &  \cellcolor{darkgreen}98 &  \cellcolor{darkgreen}38 &  \cellcolor{darkgreen}20 &  \cellcolor{darkgreen} 12 \\  
   \midrule
\parbox[t]{2mm}{\multirow{7}{*}{\rotatebox[origin=c]{90}{ViT}}}  &  None  &  \cellcolor{lightgray}False &  \cellcolor{lightgray}100 &  \cellcolor{lightgray}4 &  \cellcolor{lightgray}1 &  \cellcolor{lightgray} 0 \\  \cline{2-7}
   &  \multirow{2}{*}{JPEG}  &  \cellcolor{lightblue}False &  \cellcolor{lightblue}100 &  \cellcolor{lightblue}90 &  \cellcolor{lightblue}56 &  \cellcolor{lightblue} 26 \\  \cline{3-7}
   &     &  \cellcolor{darkblue}True &  \cellcolor{darkblue}100 &  \cellcolor{darkblue}11 &  \cellcolor{darkblue}4 &  \cellcolor{darkblue} 2 \\  \cline{2-7}
   &  \multirow{2}{*}{ELIC}  &  \cellcolor{lightred}False &  \cellcolor{lightred}99 &  \cellcolor{lightred}98 &  \cellcolor{lightred}97 &  \cellcolor{lightred} 95 \\  \cline{3-7}
   &     &  \cellcolor{darkred}True &  \cellcolor{darkred}99 &  \cellcolor{darkred}50 &  \cellcolor{darkred}20 &  \cellcolor{darkred} 9 \\  \cline{2-7}
   &  \multirow{2}{*}{HiFiC}  &  \cellcolor{lightgreen}False &  \cellcolor{lightgreen}100 &  \cellcolor{lightgreen}99 &  \cellcolor{lightgreen}98 &  \cellcolor{lightgreen} 97 \\  \cline{3-7}
   &     &  \cellcolor{darkgreen}True &  \cellcolor{darkgreen}100 &  \cellcolor{darkgreen}26 &  \cellcolor{darkgreen}8 &  \cellcolor{darkgreen} 4 \\  \bottomrule
    \end{tabular}
    \label{tab:linf_small}
\end{table}

All three defense methods showed promising results in \Cref{fig:l_inf}, showing almost no change in accuracy after all ``black-box'' attacks when using the ResNet50 model (see \Cref{fig:l_inf,fig:l_2}). For ``black-box'' attacks against ViT, iFGSM and PGD lead to a drop in accuracy at epsilon $\frac{4}{255}$ for JPEG, but still achieve a much higher accuracy than the baseline. The learned compression algorithms show much better performance for ViT, with ELIC and HiFiC achieving an accuracy comparable to before the attack even for high epsilon values.

Adversarial images crafted with the CW and DeepFool attacks were easier to defend against as shown in \Cref{fig:l_2}. This could be attributed to the fact that these attacks find minimal perturbations, compared to the other three attacks that find perturbations within the given bounds, which are not necessarily minimal. There are also more hyperparameters to tune for an optimal attack and a longer runtime. These factors lead us to focus less on these attacks and not conduct additional experiments with CW or DeepFool.

Attacking the entire pipeline drastically weakens the effect of all three defenses. There is still improvement over no defense, but the accuracy is not comparable to the experiments where the gradient information was not propagated through the defenses.

These results indicate three major things:
\begin{itemize}
    \item The ViT used is less robust against adversarial attack.
    \item The tested learned compressions perform better for ViT.
    \item Learned compression is a better defence than JPEG.
    \item Even for learned compression, the effectiveness of compression-based defenses is greatly diminished by creating adversarial images where gradient information was propagated through the defense as seen in \cite{shin2017jpeg}.
\end{itemize}
\subsubsection{ImageNet results}
\begin{figure}[t]
    \centering
    \includegraphics[width=0.99\linewidth]{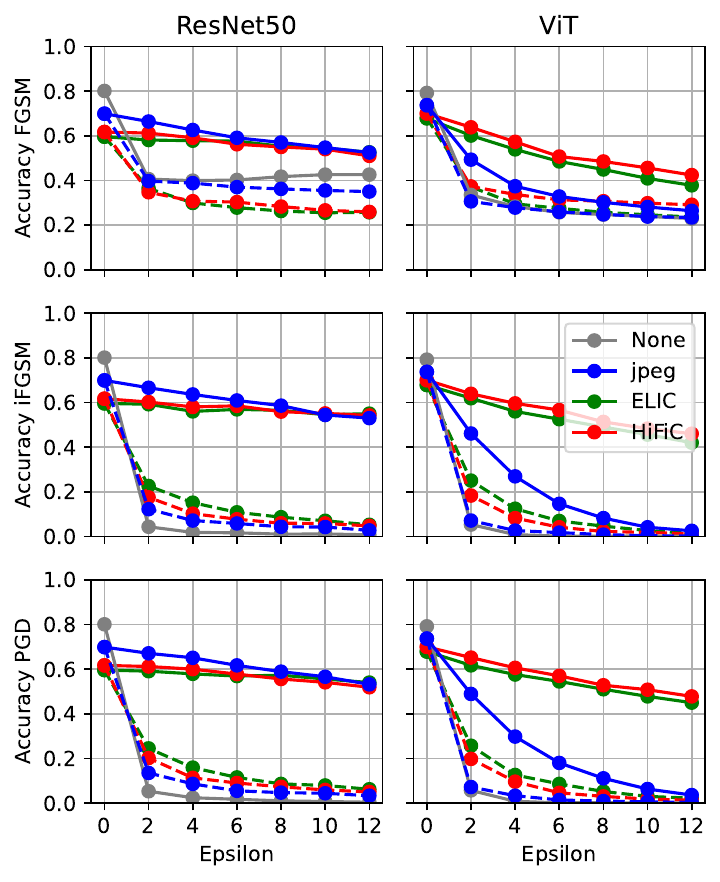}
    \caption{
     Model accuracy under FGSM (top row), iFGSM (middle row), and PGD (bottom row) adversarial attacks on 1000 randomly sampled images from the ImageNet dataset. Results demonstrate that the same defense patterns observed on Imagenette generalize to the more complex ImageNet classification task, though with lower overall accuracy due to the increased task difficulty.
    }
    \label{fig:l_inf_1000}
\end{figure}

After analyzing these results, we decided to repeat some experiments on 1000 random images taken from the ImageNet dataset, to see how the defenses perform on this harder task, with 1000 possible classes instead of 10 and a much lower baseline accuracy (both models achieve an accuracy of about 0.8 on ImageNet). The results of these experiments can be found in \Cref{fig:l_inf_1000}. In this experimental setup, all defenses show a larger accuracy decrease without an attack (epsilon$=0$). This decrease is larger for ResNet50 than for the ViT. There is also a more noticeable decrease as the epsilon constraint of the attack gets larger for all attacks. On Imagenette the accuracy was almost constant for effective defenses. However, the results on ImageNet generally show the same trends as those discussed before.

\subsection{Computational overhead}
We also computed the time it takes the model during inference to show that these defenses are feasible in practice. 
For this test, we used an Nvidia RTX3090 and ResNet50. We did not include the time the models need to initialize. Without a defense it took 20.8 seconds to classify the 3925 images in the Imagenette dataset, or 5 ms per image. 
With JPEG as a defense, it took 8 ms per image. Using ELIC and HiFiC it took 14 ms per image. Even when compressing and decompressing the images 5 times in sequence, this only increased to 33 ms for HiFiC and 36 ms for ELIC. 
These timings show that running these compression algorithms in an ML pipeline is feasible.

\subsection{Quality ablation}
This section compares different compression strengths for all the compressions used as defenses. Complete tables can be found in the Appendix, see \Cref{tab:abl_jpeg,tab:abl_hific,tab:alb_elic}.

\begin{figure}[t]
    \centering
    \includegraphics[width=1.0\linewidth]{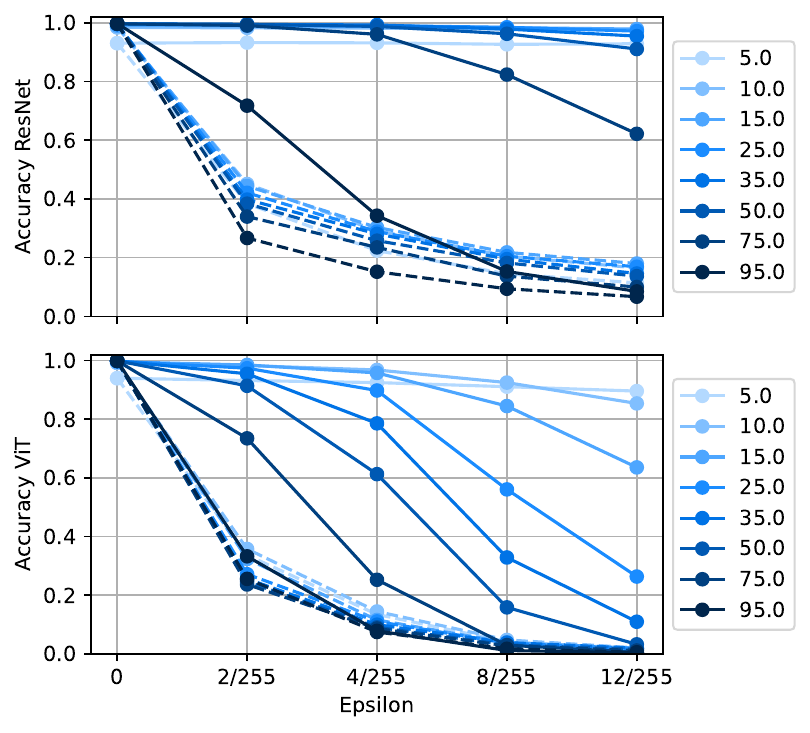}
    \caption{
    Comparison of model accuracy for different JPEG quality levels under iFGSM attacks for ResNet50 (top) and ViT (bottom). Dashed lines show results when gradient information was available to the attack (``white-box'' setting). Lower quality settings provide better defense against strong attacks but reduce clean accuracy, with quality level 25.0 offering the best trade-off.
    }
    \label{fig:abl_jpeg}
\end{figure}

\subsubsection{JPEG}
\Cref{fig:abl_jpeg} shows the accuracy for different JPEG qualities when attacked with iFGSM. For ResNet50, only the high quality levels (75.0,95.0) were vulnerable to a standard attack. When using the stronger attack, which propagates the gradients through the defense, no quality level achieved a high accuracy and therefore a successful defense. For an attack of 8/255 the levels 15.0 and 25.0 achieved the highest accuracy at $\approx 0.2$. In the ViT experiment, a larger spread of results can be observed for ``black-box'' attacks, with lower quality levels performing better at high epsilon values. JPEG with a quality level 5.0 showed a large decrease in accuracy without any attack. For ``white-box'' attacks none of the levels provide almost any defence. Considering these observations, we decided to use a quality level of 25.0, as this achieves a good performance, does not degrade the baseline accuracy and has been used in previous work by Shin \etal \cite{shin2017jpeg}.

\subsubsection{HiFiC}
\begin{figure}[t]
    \centering
    \includegraphics[width=1.0\linewidth]{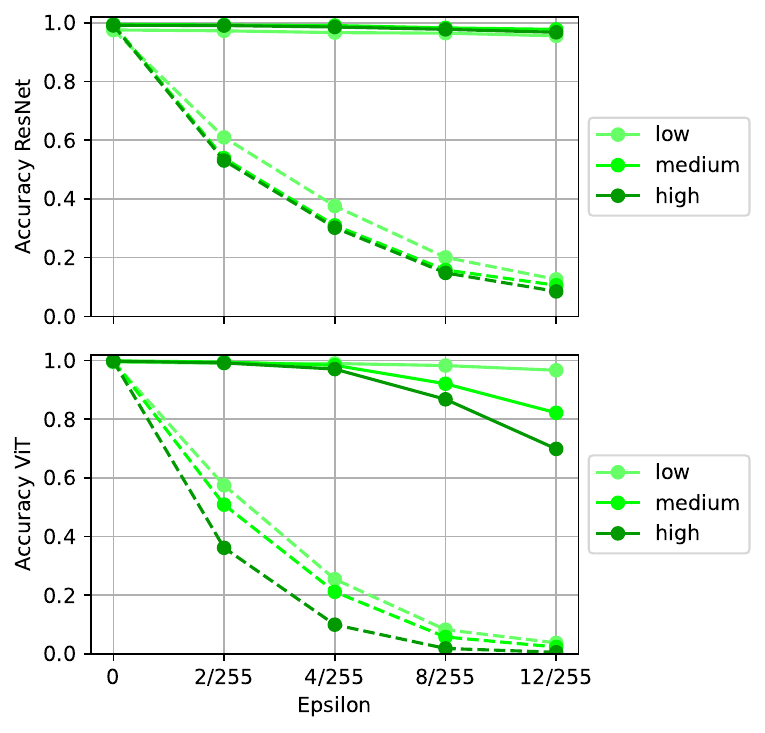}
    \caption{
    Comparison of model accuracy for different HiFiC quality settings (low, medium, high) under iFGSM attacks for ResNet50 (top) and ViT (bottom). HiFiC low provides the strongest defense with minimal impact on clean accuracy, particularly for the ViT model.
    }
    \label{fig:abl_hific}
\end{figure}

\Cref{fig:abl_hific} shows that all the different compression strengths worked well as a defense for ResNet50, but only HiFiC low achieved a high accuracy for the ViT, with the other two qualities showing a decrease in accuracy for epsilon larger than $ \frac{4}{255}$. All levels were very vulnerable when the gradient was passed through the defense. Since the lowest quality only decreased baseline performance a little, we used HiFiC low for the main experiments.

\subsubsection{ELIC}
\begin{figure}[ht]
    \centering
    \includegraphics[width=1.0\linewidth]{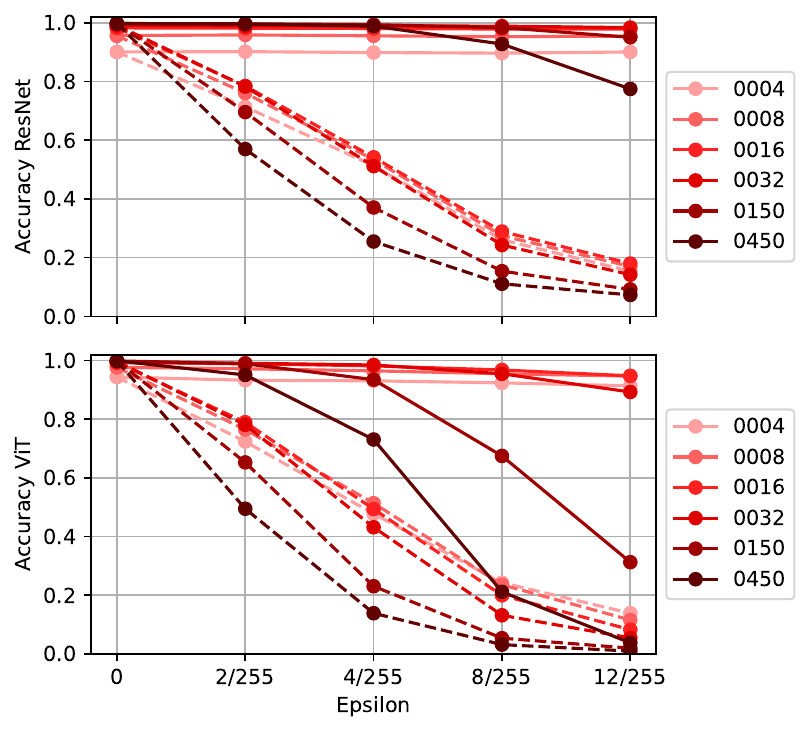}
    \caption{
    Comparison of model accuracy for different ELIC quality parameters (0004 through 0450) under iFGSM attacks for ResNet50 (top) and ViT (bottom). ELIC 0016 provides an optimal balance between defense strength and clean image accuracy, with comparable bitrate to HiFiC low.
    }
    \label{fig:abl_elic}
\end{figure}

\Cref{fig:abl_elic} shows the results for ELIC. The higher quality levels show decreased performance for high epsilon values. The lowest quality levels show a decreased accuracy without any attack. There is a larger difference between quality levels when the gradient is propagated through the defense compared to HiFiC. For the main experiment we decided to use ELIC 0016, as it achieved good accuracies without visibly decreasing the baseline performance, and because it is similar in BPP to HiFiC low. 

\subsection{Sequential defense}
\subsubsection{Imagenette}
\begin{figure}[t]
    \centering
    \includegraphics[width=1.0\linewidth]{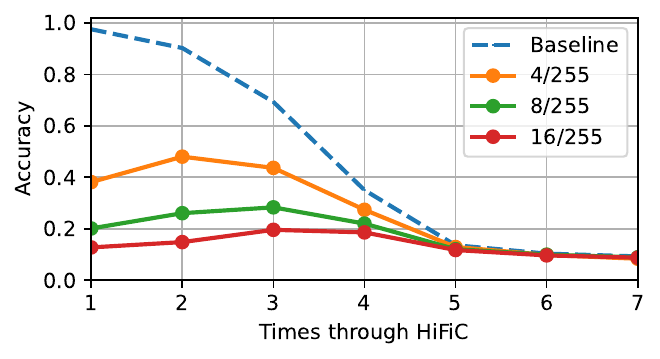}
    \caption{
    Defense effectiveness of sequential HiFiC low compression against iFGSM attacks on Imagenette using ResNet50. Each point represents N consecutive compression/decompression cycles. The baseline accuracy rapidly decreases after multiple iterations, limiting the practical utility of sequential HiFiC defenses.
    }
    \label{fig:hific_seq_low}
\end{figure}

\begin{figure}[t]
    \centering
    \includegraphics[width=1.0\linewidth]{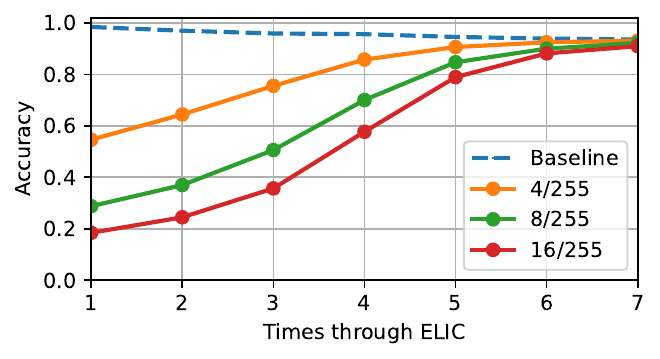}
    \caption{
    Defense effectiveness of sequential ELIC 0016 compression against iFGSM attacks on Imagenette using ResNet50. Accuracy against adversarial examples improves with multiple compression cycles while maintaining reasonable baseline accuracy, demonstrating better sequential defense properties than HiFiC.
    }
    \label{fig:elic_seq_low}
\end{figure}

\begin{figure}[t]
    \centering
    \includegraphics[width=0.9\linewidth]{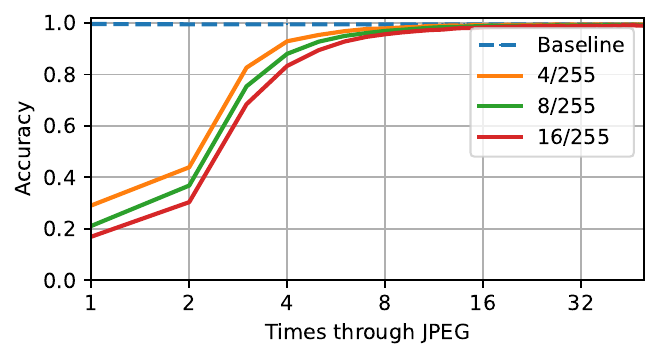}
    \includegraphics[width=0.9\linewidth]{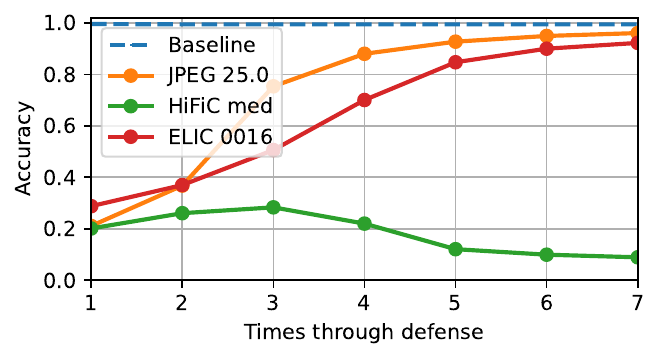}
    \caption{
    Comparative analysis of sequential defense performance across compression methods on Imagenette with iFGSM attacks. Top: JPEG defense at different iteration counts. Bottom: Comparison between JPEG, HiFiC, and ELIC for epsilon 8/255. JPEG achieves the fastest convergence toward baseline accuracy with minimal clean accuracy degradation.
    }
    \label{fig:seq-jpeg_large}
\end{figure}

For JPEG, we used quality level 25.0 for the results shown in \Cref{fig:seq-jpeg_large,fig:seq-jpeg_small}. For HiFiC and ELIC, there is a tradeoff between a decrease in accuracy without an attack on low qualities and a decrease in effectiveness on high qualities. The baseline accuracy using HiFiC low deteriorates quickly, making it unusable as a sequential defense, see \Cref{fig:hific_seq_low}. ELIC 0016 also decreases the baseline accuracy, but it was much slower than HiFiC low. This leads to a promising defense that reaches close to baseline accuracy with seven sequential defense iterations, as seen in \Cref{fig:elic_seq_low}.

The experiments indicate a clear trend, showing that running an image through a defense multiple times increases its effectiveness for all defenses. JPEG showed the fastest increase, achieving an accuracy of over 0.9 for epsilon $\frac{8}{255}$ after 5 iterations and converging towards the baseline. The decrease in baseline accuracy for JPEG is negligible even after 50 iterations. HiFiC and ELIC also show an increased accuracy for each additional sequential iteration, achieving accuracies of about 0.4 with 7 iterations. Because of the large amount of gradient information, we were unable to compute results for more than 7 iterations of ELIC or HiFiC.

\Cref{fig:images} shows how the image quality deteriorates when compressing and decompressing multiple times in sequence. Both of the learned compressions show a stark difference to the original image. ELIC introduces black/red artifacts which take up parts of the image. The lowest quality of HiFiC leads to a much brighter image, which lost most of the color information. JPEG  performed best, while it also introduces artifacts and blurs the image a lot, the image still looks similar after multiple passes through the compression.

\subsubsection{ImageNet}
\begin{figure}[t]
    \centering
    \begin{subfigure}[b]{0.9\linewidth}
        \centering
        \includegraphics[width=1.0\linewidth]{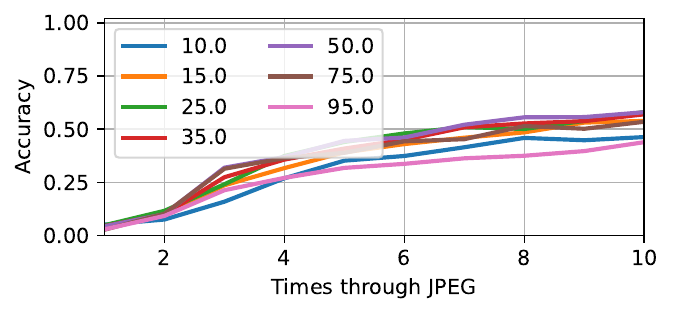}
        \label{fig:seq_jpeg_inet_resnet}
    \end{subfigure}
    
    \begin{subfigure}[b]{0.9\linewidth}
        \centering
        \includegraphics[width=1.0\linewidth]{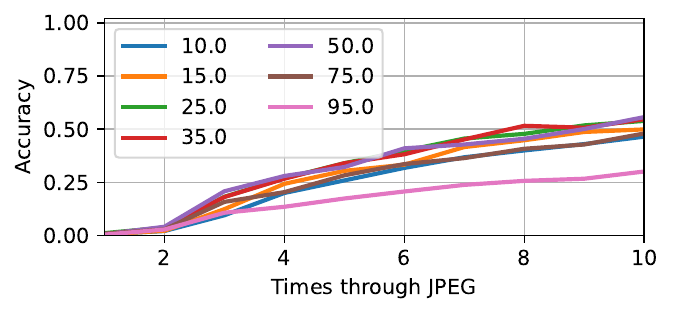}
        \label{fig:seq_jpeg_inet_vit}
    \end{subfigure}
    
    \caption{
    Comparison of sequential JPEG defense effectiveness against iFGSM attacks at various quality levels on the ImageNet dataset for ResNet50 (top) and ViT (bottom). Lower quality settings show decreased clean accuracy but provide stronger defenses against adversarial examples.
    }
    \label{fig:seq_jpeg_inet}
\end{figure}

\begin{figure}[t]
    \centering
    \includegraphics[width=0.9\linewidth]{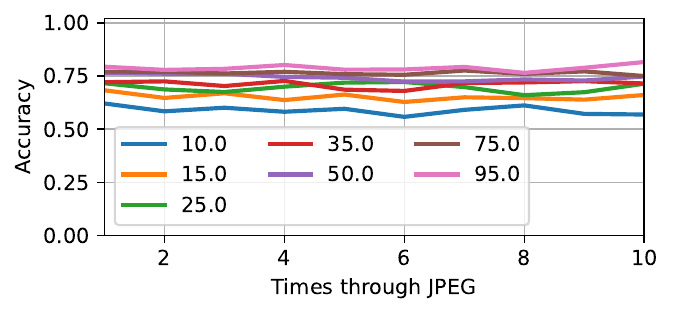}
    \caption{
    Baseline accuracy (no attack) after JPEG compression cycles at different quality levels on ImageNet using ResNet50. The plot demonstrates how image quality degradation affects classification performance even without adversarial perturbations.
    }
    \label{fig:seq-jpeg_inetbase}
\end{figure}

Experiments in \Cref{fig:seq_jpeg_inet} with sequential JPEG defense on ImageNet yielded similar patterns but with lower overall accuracy, partly due to decreased baseline performance. Lower quality levels reduced clean image accuracy, with this reduction primarily dependent on the quality parameter rather than iteration count, as accuracy remained relatively stable across multiple compression cycles.

The lines' jaggedness compared to earlier could come from the random sampling.
However, even considering the lower baselines, no quality seems to converge towards the baseline with 10 sequential passes. This indicates a deeper sequential defense is needed for this more challenging task.

\begin{figure}[t]
    \centering
    \includegraphics[width=0.9\linewidth]{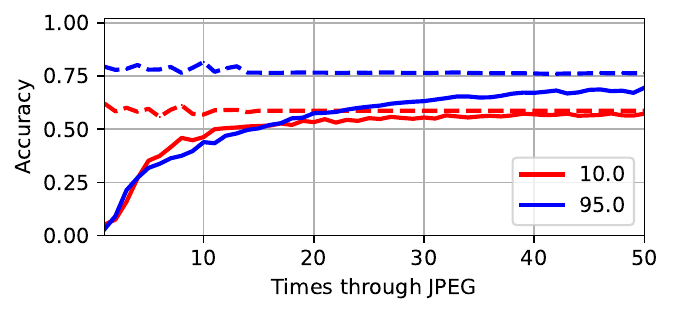}
    \caption{
    Extended sequential JPEG defense for quality levels 10.0 and 95.0 up to 50 iterations on ImageNet. Quality 10.0 converges faster but to a lower accuracy ceiling, while quality 95.0 continues to improve even after 50 iterations but at a slower rate.
    }
    \label{fig:seq_jpeg_deep}
\end{figure}

An experiment with a deeper sequential JPEG defense, see \Cref{fig:seq_jpeg_deep}, shows this, as quality 10.0 seems to converge. For quality 95.0 the accuracy still improves even at 50 iterations. The jaggedness of the baseline for low iterations is due to using different subsets of ImageNet. This was changed later in the experiment for more consistent results.

\section{Conclusion}
\label{sec:conclusion}
This paper demonstrates that human-aligned learned compression can effectively defend against adversarial attacks.
We show that HiFiC and ELIC have advantages over JPEG, as they do not significantly decrease the baseline accuracy of an image classification model even at low BPP. 
However, we also show the weaknesses of such defenses in a white-box setting, as a gradient can either be directly computed or approximated, decreasing the defense's effectiveness. 
Sequential compression significantly enhances defense effectiveness, with JPEG showing the most practical balance between robustness and image quality over multiple iterations. 

There are some limitations, as we only experiment with gradient-based attacks. In real world settings, there are more possible ways to attack a model such as gradient free attacks \cite{uesato2018adversarialriskdangersevaluating,engstrom2019exploringlandscapespatialrobustness,gilmer2018motivatingrulesgameadversarial}. Further work could include results for these defenses in settings that include these additional threats and combines image compression with other defensive measures to create more robust deep learning models.


While these defenses are weakened in white-box settings, they offer meaningful protection. Future work should explore combining compression-based defenses with other techniques and test against other threats to develop more robust systems that align with human visual perception.

\clearpage
{
    \small
    \bibliographystyle{ieeenat_fullname}
    \bibliography{main}

\begin{thebibliography}{31}
\providecommand{\natexlab}[1]{#1}
\providecommand{\url}[1]{\texttt{#1}}
\expandafter\ifx\csname urlstyle\endcsname\relax
  \providecommand{\doi}[1]{doi: #1}\else
  \providecommand{\doi}{doi: \begingroup \urlstyle{rm}\Url}\fi

\bibitem[Agustsson et~al.(2019)Agustsson, Tschannen, Mentzer, Timofte, and Gool]{agustsson2019generativeadversarialnetworksextreme}
Eirikur Agustsson, Michael Tschannen, Fabian Mentzer, Radu Timofte, and Luc~Van Gool.
\newblock Generative adversarial networks for extreme learned image compression.
\newblock In \emph{Proceedings of the IEEE/CVF International Conference on Computer Vision (ICCV)}, 2019.

\bibitem[Carlini and Wagner(2017)]{carlini2017evaluatingrobustnessneuralnetworks}
Nicholas Carlini and David Wagner.
\newblock Towards evaluating the robustness of neural networks, 2017.

\bibitem[Cheng et~al.(2020)Cheng, Sun, Takeuchi, and Katto]{cheng2020learnedimagecompressiondiscretized}
Zhengxue Cheng, Heming Sun, Masaru Takeuchi, and Jiro Katto.
\newblock Learned image compression with discretized gaussian mixture likelihoods and attention modules, 2020.

\bibitem[Deng et~al.(2009)Deng, Dong, Socher, Li, Li, and Fei-Fei]{deng2009imagenet}
Jia Deng, Wei Dong, Richard Socher, Li-Jia Li, Kai Li, and Li Fei-Fei.
\newblock Imagenet: A large-scale hierarchical image database.
\newblock In \emph{2009 IEEE conference on computer vision and pattern recognition}, pages 248--255. Ieee, 2009.

\bibitem[Dosovitskiy et~al.(2021)Dosovitskiy, Beyer, Kolesnikov, Weissenborn, Zhai, Unterthiner, Dehghani, Minderer, Heigold, Gelly, Uszkoreit, and Houlsby]{dosovitskiy2021imageworth16x16words}
Alexey Dosovitskiy, Lucas Beyer, Alexander Kolesnikov, Dirk Weissenborn, Xiaohua Zhai, Thomas Unterthiner, Mostafa Dehghani, Matthias Minderer, Georg Heigold, Sylvain Gelly, Jakob Uszkoreit, and Neil Houlsby.
\newblock An image is worth 16x16 words: Transformers for image recognition at scale, 2021.

\bibitem[Dziugaite et~al.(2016)Dziugaite, Ghahramani, and Roy]{dziugaite2016studyeffectjpgcompression}
Gintare~Karolina Dziugaite, Zoubin Ghahramani, and Daniel~M. Roy.
\newblock A study of the effect of jpg compression on adversarial images, 2016.

\bibitem[Engstrom et~al.(2019)Engstrom, Tran, Tsipras, Schmidt, and Madry]{engstrom2019exploringlandscapespatialrobustness}
Logan Engstrom, Brandon Tran, Dimitris Tsipras, Ludwig Schmidt, and Aleksander Madry.
\newblock Exploring the landscape of spatial robustness, 2019.

\bibitem[Galteri et~al.(2019)Galteri, Seidenari, Bertini, and Del~Bimbo]{galteri2019deep}
Leonardo Galteri, Lorenzo Seidenari, Marco Bertini, and Alberto Del~Bimbo.
\newblock Deep universal generative adversarial compression artifact removal.
\newblock \emph{IEEE Transactions on Multimedia}, 21\penalty0 (8):\penalty0 2131--2145, 2019.

\bibitem[Gilmer et~al.(2018)Gilmer, Adams, Goodfellow, Andersen, and Dahl]{gilmer2018motivatingrulesgameadversarial}
Justin Gilmer, Ryan~P. Adams, Ian Goodfellow, David Andersen, and George~E. Dahl.
\newblock Motivating the rules of the game for adversarial example research, 2018.

\bibitem[Goodfellow et~al.(2014)Goodfellow, Shlens, and Szegedy]{goodfellow2014explaining}
Ian~J Goodfellow, Jonathon Shlens, and Christian Szegedy.
\newblock Explaining and harnessing adversarial examples.
\newblock \emph{arXiv preprint arXiv:1412.6572}, 2014.

\bibitem[He et~al.(2021)He, Zheng, Sun, Wang, and Qin]{He_2021_CVPR}
Dailan He, Yaoyan Zheng, Baocheng Sun, Yan Wang, and Hongwei Qin.
\newblock Checkerboard context model for efficient learned image compression.
\newblock In \emph{Proceedings of the IEEE/CVF Conference on Computer Vision and Pattern Recognition (CVPR)}, pages 14771--14780, 2021.

\bibitem[He et~al.(2022)He, Yang, Peng, Ma, Qin, and Wang]{he2022elicefficientlearnedimage}
Dailan He, Ziming Yang, Weikun Peng, Rui Ma, Hongwei Qin, and Yan Wang.
\newblock Elic: Efficient learned image compression with unevenly grouped space-channel contextual adaptive coding, 2022.

\bibitem[He et~al.(2015)He, Zhang, Ren, and Sun]{he2015deepresiduallearningimage}
Kaiming He, Xiangyu Zhang, Shaoqing Ren, and Jian Sun.
\newblock Deep residual learning for image recognition, 2015.

\bibitem[Hinton et~al.(2015)Hinton, Vinyals, and Dean]{hinton2015distillingknowledgeneuralnetwork}
Geoffrey Hinton, Oriol Vinyals, and Jeff Dean.
\newblock Distilling the knowledge in a neural network, 2015.

\bibitem[Ilyas et~al.(2019)Ilyas, Santurkar, Tsipras, Engstrom, Tran, and Madry]{ilyas2019adversarial}
Andrew Ilyas, Shibani Santurkar, Dimitris Tsipras, Logan Engstrom, Brandon Tran, and Aleksander Madry.
\newblock Adversarial examples are not bugs, they are features.
\newblock \emph{Advances in neural information processing systems}, 32, 2019.

\bibitem[Kim(2020)]{kim2020torchattacks}
Hoki Kim.
\newblock Torchattacks: A pytorch repository for adversarial attacks.
\newblock \emph{arXiv preprint arXiv:2010.01950}, 2020.

\bibitem[Kurakin et~al.(2017)Kurakin, Goodfellow, and Bengio]{kurakin2017adversarialexamplesphysicalworld}
Alexey Kurakin, Ian Goodfellow, and Samy Bengio.
\newblock Adversarial examples in the physical world, 2017.

\bibitem[Liu et~al.(2023)Liu, Sun, and Katto]{LiU_2023_CVPR}
Jinming Liu, Heming Sun, and Jiro Katto.
\newblock Learned image compression with mixed transformer-cnn architectures.
\newblock In \emph{Proceedings of the IEEE/CVF Conference on Computer Vision and Pattern Recognition (CVPR)}, pages 14388--14397, 2023.

\bibitem[Madry et~al.(2019)Madry, Makelov, Schmidt, Tsipras, and Vladu]{madry2019deeplearningmodelsresistant}
Aleksander Madry, Aleksandar Makelov, Ludwig Schmidt, Dimitris Tsipras, and Adrian Vladu.
\newblock Towards deep learning models resistant to adversarial attacks, 2019.

\bibitem[Mentzer et~al.(2020)Mentzer, Toderici, Tschannen, and Agustsson]{mentzer2020highfidelitygenerativeimagecompression}
Fabian Mentzer, George Toderici, Michael Tschannen, and Eirikur Agustsson.
\newblock High-fidelity generative image compression, 2020.

\bibitem[Minnen et~al.(2018)Minnen, Ball\'{e}, and Toderici]{minnen2018jointautoregressivehierarchicalpriors}
David Minnen, Johannes Ball\'{e}, and George~D Toderici.
\newblock Joint autoregressive and hierarchical priors for learned image compression.
\newblock In \emph{Advances in Neural Information Processing Systems}. Curran Associates, Inc., 2018.

\bibitem[Moosavi-Dezfooli et~al.(2016)Moosavi-Dezfooli, Fawzi, and Frossard]{moosavidezfooli2016deepfoolsimpleaccuratemethod}
Seyed-Mohsen Moosavi-Dezfooli, Alhussein Fawzi, and Pascal Frossard.
\newblock Deepfool: a simple and accurate method to fool deep neural networks, 2016.

\bibitem[Papernot et~al.(2016)Papernot, McDaniel, Wu, Jha, and Swami]{papernot2016distillationdefenseadversarialperturbations}
Nicolas Papernot, Patrick McDaniel, Xi Wu, Somesh Jha, and Ananthram Swami.
\newblock Distillation as a defense to adversarial perturbations against deep neural networks, 2016.

\bibitem[Paszke et~al.(2019)Paszke, Gross, Massa, Lerer, Bradbury, Chanan, Killeen, Lin, Gimelshein, Antiga, Desmaison, Köpf, Yang, DeVito, Raison, Tejani, Chilamkurthy, Steiner, Fang, Bai, and Chintala]{paszke2019pytorchimperativestylehighperformance}
Adam Paszke, Sam Gross, Francisco Massa, Adam Lerer, James Bradbury, Gregory Chanan, Trevor Killeen, Zeming Lin, Natalia Gimelshein, Luca Antiga, Alban Desmaison, Andreas Köpf, Edward Yang, Zach DeVito, Martin Raison, Alykhan Tejani, Sasank Chilamkurthy, Benoit Steiner, Lu Fang, Junjie Bai, and Soumith Chintala.
\newblock Pytorch: An imperative style, high-performance deep learning library, 2019.

\bibitem[Riba et~al.(2020)Riba, Mishkin, Ponsa, Rublee, and Bradski]{eriba2019kornia}
Edgar Riba, Dmytro Mishkin, Daniel Ponsa, Ethan Rublee, and Gary Bradski.
\newblock Kornia: an open source differentiable computer vision library for pytorch.
\newblock In \emph{Proceedings of the IEEE/CVF Winter Conference on Applications of Computer Vision}, pages 3674--3683, 2020.

\bibitem[Shen et~al.(2017)Shen, Wu, and Suk]{shenmedical2017}
Dinggang Shen, Guorong Wu, and Heung-Il Suk.
\newblock Deep learning in medical image analysis.
\newblock \emph{Annual Review of Biomedical Engineering}, 19\penalty0 (Volume 19, 2017):\penalty0 221--248, 2017.

\bibitem[Shin and Song(2017)]{shin2017jpeg}
Richard Shin and Dawn Song.
\newblock Jpeg-resistant adversarial images.
\newblock In \emph{NIPS 2017 workshop on machine learning and computer security}, page~8, 2017.

\bibitem[Szegedy et~al.(2013)Szegedy, Toshev, and Erhan]{NIPS2013_f7cade80}
Christian Szegedy, Alexander Toshev, and Dumitru Erhan.
\newblock Deep neural networks for object detection.
\newblock In \emph{Advances in Neural Information Processing Systems}. Curran Associates, Inc., 2013.

\bibitem[Szegedy et~al.(2014)Szegedy, Zaremba, Sutskever, Bruna, Erhan, Goodfellow, and Fergus]{szegedy2014intriguingpropertiesneuralnetworks}
Christian Szegedy, Wojciech Zaremba, Ilya Sutskever, Joan Bruna, Dumitru Erhan, Ian Goodfellow, and Rob Fergus.
\newblock Intriguing properties of neural networks, 2014.

\bibitem[Uesato et~al.(2018)Uesato, O'Donoghue, van~den Oord, and Kohli]{uesato2018adversarialriskdangersevaluating}
Jonathan Uesato, Brendan O'Donoghue, Aaron van~den Oord, and Pushmeet Kohli.
\newblock Adversarial risk and the dangers of evaluating against weak attacks, 2018.

\bibitem[Zou et~al.(2023)Zou, Chen, Shi, Guo, and Ye]{10028728}
Zhengxia Zou, Keyan Chen, Zhenwei Shi, Yuhong Guo, and Jieping Ye.
\newblock Object detection in 20 years: A survey.
\newblock \emph{Proceedings of the IEEE}, 111\penalty0 (3):\penalty0 257--276, 2023.

\end{thebibliography}
}
\appendix
\label{sec:appendix}
\onecolumn
\section{Tables}
\FloatBarrier
\begin{table}[H]
    \centering
    \begin{tabular}{|c|c|}
    \hline
         Attack& Hyperparameters \\ \hline
         FGSM& eps=epsilon  \\ \hline
         iFGSM& eps=epsilon,alpha=epsilon/4, steps = 10 \\ \hline
         PGD& eps=epsilon, alpha=epsilon/4,steps=10,randomstart=True  \\ \hline
         CW&  c=1, kappa=0, steps=50, lr=0.01\\ \hline
         DeepFool& steps=50, overshoot=0.02  \\ \hline
    \end{tabular}
    \caption{
    Hyperparameters used for each adversarial attack method. This table details the specific configuration for the experiments' FGSM, iFGSM, PGD, CW, and DeepFool attacks.
    }
    \label{tab:hyperparams}
\end{table}

\label{sec:table_l2}
\begin{table}
    \centering
    \begin{tabular}{|c|c|c|c|c|c|c|c|c|c|c|c|}
        \hline 
        \multicolumn{5}{|c|}{Attack parameters} & \multicolumn{7}{c|}{$l_2$ norm value}\\
        \hline
        Attack & Model & Defense & Through & Baseline& 4&5&6&7&8&16&32\\
        \hline
        \multirow{14}{*}{CW}  &  \multirow{7}{*}{ResNet50}  &  None  &  \cellcolor{lightgray}False &  \cellcolor{lightgray}0.998 &  \cellcolor{lightgray}0.944 &  \cellcolor{lightgray}0.787 &  \cellcolor{lightgray}0.529 &  \cellcolor{lightgray}0.304 &  \cellcolor{lightgray}0.193 &  \cellcolor{lightgray}0.054 &  \cellcolor{lightgray} 0.051 \\  \cline{3-12}
   &     &  \multirow{2}{*}{jpeg}  &  \cellcolor{lightblue}False &  \cellcolor{lightblue}0.996 &  \cellcolor{lightblue}0.996 &  \cellcolor{lightblue}0.996 &  \cellcolor{lightblue}0.995 &  \cellcolor{lightblue}0.996 &  \cellcolor{lightblue}0.996 &  \cellcolor{lightblue}0.996 &  \cellcolor{lightblue} 0.996 \\  \cline{4-12}
   &     &     &  \cellcolor{darkblue}True &  \cellcolor{darkblue}0.996 &  \cellcolor{darkblue}0.518 &  \cellcolor{darkblue}0.338 &  \cellcolor{darkblue}0.285 &  \cellcolor{darkblue}0.281 &  \cellcolor{darkblue}0.281 &  \cellcolor{darkblue}0.281 &  \cellcolor{darkblue} 0.281 \\  \cline{3-12}
   &     &  \multirow{2}{*}{ELIC}  &  \cellcolor{lightred}False &  \cellcolor{lightred}0.998 &  \cellcolor{lightred}0.998 &  \cellcolor{lightred}0.998 &  \cellcolor{lightred}0.998 &  \cellcolor{lightred}0.997 &  \cellcolor{lightred}0.997 &  \cellcolor{lightred}0.997 &  \cellcolor{lightred} 0.997 \\  \cline{4-12}
   &     &     &  \cellcolor{darkred}True &  \cellcolor{darkred}0.998 &  \cellcolor{darkred}0.958 &  \cellcolor{darkred}0.869 &  \cellcolor{darkred}0.724 &  \cellcolor{darkred}0.554 &  \cellcolor{darkred}0.429 &  \cellcolor{darkred}0.225 &  \cellcolor{darkred} 0.225 \\  \cline{3-12}
   &     &  \multirow{2}{*}{HiFiC}  &  \cellcolor{lightgreen}False &  \cellcolor{lightgreen}0.975 &  \cellcolor{lightgreen}0.975 &  \cellcolor{lightgreen}0.975 &  \cellcolor{lightgreen}0.975 &  \cellcolor{lightgreen}0.975 &  \cellcolor{lightgreen}0.975 &  \cellcolor{lightgreen}0.975 &  \cellcolor{lightgreen} 0.975 \\  \cline{4-12}
   &     &     &  \cellcolor{darkgreen}True &  \cellcolor{darkgreen}0.975 &  \cellcolor{darkgreen}0.972 &  \cellcolor{darkgreen}0.961 &  \cellcolor{darkgreen}0.932 &  \cellcolor{darkgreen}0.9 &  \cellcolor{darkgreen}0.868 &  \cellcolor{darkgreen}0.596 &  \cellcolor{darkgreen} 0.596 \\  \cline{2-12}
   &  \multirow{7}{*}{ViT}  &  None  &  \cellcolor{lightgray}False &  \cellcolor{lightgray}0.999 &  \cellcolor{lightgray}0.803 &  \cellcolor{lightgray}0.494 &  \cellcolor{lightgray}0.231 &  \cellcolor{lightgray}0.097 &  \cellcolor{lightgray}0.045 &  \cellcolor{lightgray}0.005 &  \cellcolor{lightgray} 0.005 \\  \cline{3-12}
   &     &  \multirow{2}{*}{jpeg}  &  \cellcolor{lightblue}False &  \cellcolor{lightblue}0.998 &  \cellcolor{lightblue}0.997 &  \cellcolor{lightblue}0.995 &  \cellcolor{lightblue}0.995 &  \cellcolor{lightblue}0.995 &  \cellcolor{lightblue}0.995 &  \cellcolor{lightblue}0.995 &  \cellcolor{lightblue} 0.995 \\  \cline{4-12}
   &     &     &  \cellcolor{darkblue}True &  \cellcolor{darkblue}0.998 &  \cellcolor{darkblue}0.409 &  \cellcolor{darkblue}0.287 &  \cellcolor{darkblue}0.267 &  \cellcolor{darkblue}0.264 &  \cellcolor{darkblue}0.264 &  \cellcolor{darkblue}0.264 &  \cellcolor{darkblue} 0.264 \\  \cline{3-12}
   &     &  \multirow{2}{*}{ELIC}  &  \cellcolor{lightred}False &  \cellcolor{lightred}0.999 &  \cellcolor{lightred}0.999 &  \cellcolor{lightred}0.999 &  \cellcolor{lightred}0.999 &  \cellcolor{lightred}0.999 &  \cellcolor{lightred}0.999 &  \cellcolor{lightred}0.999 &  \cellcolor{lightred} 0.999 \\  \cline{4-12}
   &     &     &  \cellcolor{darkred}True &  \cellcolor{darkred}0.999 &  \cellcolor{darkred}0.934 &  \cellcolor{darkred}0.796 &  \cellcolor{darkred}0.555 &  \cellcolor{darkred}0.346 &  \cellcolor{darkred}0.228 &  \cellcolor{darkred}0.137 &  \cellcolor{darkred} 0.137 \\  \cline{3-12}
   &     &  \multirow{2}{*}{HiFiC}  &  \cellcolor{lightgreen}False &  \cellcolor{lightgreen}0.995 &  \cellcolor{lightgreen}0.995 &  \cellcolor{lightgreen}0.995 &  \cellcolor{lightgreen}0.995 &  \cellcolor{lightgreen}0.995 &  \cellcolor{lightgreen}0.995 &  \cellcolor{lightgreen}0.995 &  \cellcolor{lightgreen} 0.995 \\  \cline{4-12}
   &     &     &  \cellcolor{darkgreen}True &  \cellcolor{darkgreen}0.995 &  \cellcolor{darkgreen}0.984 &  \cellcolor{darkgreen}0.964 &  \cellcolor{darkgreen}0.934 &  \cellcolor{darkgreen}0.888 &  \cellcolor{darkgreen}0.832 &  \cellcolor{darkgreen}0.547 &  \cellcolor{darkgreen} 0.547 \\  \hline
\multirow{14}{*}{DeepFool}  &  \multirow{7}{*}{ResNet50}  &  None  &  \cellcolor{lightgray}False &  \cellcolor{lightgray}0.998 &  \cellcolor{lightgray}0.541 &  \cellcolor{lightgray}0.493 &  \cellcolor{lightgray}0.464 &  \cellcolor{lightgray}0.452 &  \cellcolor{lightgray}0.447 &  \cellcolor{lightgray}0.443 &  \cellcolor{lightgray} 0.443 \\  \cline{3-12}
   &     &  \multirow{2}{*}{jpeg}  &  \cellcolor{lightblue}False &  \cellcolor{lightblue}0.996 &  \cellcolor{lightblue}0.995 &  \cellcolor{lightblue}0.995 &  \cellcolor{lightblue}0.995 &  \cellcolor{lightblue}0.995 &  \cellcolor{lightblue}0.995 &  \cellcolor{lightblue}0.995 &  \cellcolor{lightblue} 0.995 \\  \cline{4-12}
   &     &     &  \cellcolor{darkblue}True &  \cellcolor{darkblue}0.996 &  \cellcolor{darkblue}0.892 &  \cellcolor{darkblue}0.849 &  \cellcolor{darkblue}0.8 &  \cellcolor{darkblue}0.753 &  \cellcolor{darkblue}0.708 &  \cellcolor{darkblue}0.406 &  \cellcolor{darkblue} 0.155 \\  \cline{3-12}
   &     &  \multirow{2}{*}{ELIC}  &  \cellcolor{lightred}False &  \cellcolor{lightred}0.998 &  \cellcolor{lightred}0.998 &  \cellcolor{lightred}0.998 &  \cellcolor{lightred}0.998 &  \cellcolor{lightred}0.998 &  \cellcolor{lightred}0.998 &  \cellcolor{lightred}0.998 &  \cellcolor{lightred} 0.998 \\  \cline{4-12}
   &     &     &  \cellcolor{darkred}True &  \cellcolor{darkred}0.998 &  \cellcolor{darkred}0.917 &  \cellcolor{darkred}0.876 &  \cellcolor{darkred}0.83 &  \cellcolor{darkred}0.78 &  \cellcolor{darkred}0.738 &  \cellcolor{darkred}0.559 &  \cellcolor{darkred} 0.541 \\  \cline{3-12}
   &     &  \multirow{2}{*}{HiFiC}  &  \cellcolor{lightgreen}False &  \cellcolor{lightgreen}0.975 &  \cellcolor{lightgreen}0.975 &  \cellcolor{lightgreen}0.975 &  \cellcolor{lightgreen}0.975 &  \cellcolor{lightgreen}0.975 &  \cellcolor{lightgreen}0.975 &  \cellcolor{lightgreen}0.975 &  \cellcolor{lightgreen} 0.975 \\  \cline{4-12}
   &     &     &  \cellcolor{darkgreen}True &  \cellcolor{darkgreen}0.975 &  \cellcolor{darkgreen}0.797 &  \cellcolor{darkgreen}0.743 &  \cellcolor{darkgreen}0.69 &  \cellcolor{darkgreen}0.638 &  \cellcolor{darkgreen}0.59 &  \cellcolor{darkgreen}0.356 &  \cellcolor{darkgreen} 0.281 \\  \cline{2-12}
   &  \multirow{7}{*}{ViT}  &  None  &  \cellcolor{lightgray}False &  \cellcolor{lightgray}0.999 &  \cellcolor{lightgray}0.752 &  \cellcolor{lightgray}0.641 &  \cellcolor{lightgray}0.539 &  \cellcolor{lightgray}0.448 &  \cellcolor{lightgray}0.376 &  \cellcolor{lightgray}0.194 &  \cellcolor{lightgray} 0.171 \\  \cline{3-12}
   &     &  \multirow{2}{*}{jpeg}  &  \cellcolor{lightblue}False &  \cellcolor{lightblue}0.998 &  \cellcolor{lightblue}0.996 &  \cellcolor{lightblue}0.996 &  \cellcolor{lightblue}0.996 &  \cellcolor{lightblue}0.996 &  \cellcolor{lightblue}0.996 &  \cellcolor{lightblue}0.996 &  \cellcolor{lightblue} 0.996 \\  \cline{4-12}
   &     &     &  \cellcolor{darkblue}True &  \cellcolor{darkblue}0.998 &  \cellcolor{darkblue}0.969 &  \cellcolor{darkblue}0.95 &  \cellcolor{darkblue}0.931 &  \cellcolor{darkblue}0.912 &  \cellcolor{darkblue}0.885 &  \cellcolor{darkblue}0.684 &  \cellcolor{darkblue} 0.381 \\  \cline{3-12}
   &     &  \multirow{2}{*}{ELIC}  &  \cellcolor{lightred}False &  \cellcolor{lightred}0.999 &  \cellcolor{lightred}0.999 &  \cellcolor{lightred}0.998 &  \cellcolor{lightred}0.998 &  \cellcolor{lightred}0.998 &  \cellcolor{lightred}0.998 &  \cellcolor{lightred}0.998 &  \cellcolor{lightred} 0.998 \\  \cline{4-12}
   &     &     &  \cellcolor{darkred}True &  \cellcolor{darkred}0.999 &  \cellcolor{darkred}0.955 &  \cellcolor{darkred}0.926 &  \cellcolor{darkred}0.891 &  \cellcolor{darkred}0.854 &  \cellcolor{darkred}0.81 &  \cellcolor{darkred}0.559 &  \cellcolor{darkred} 0.417 \\  \cline{3-12}
   &     &  \multirow{2}{*}{HiFiC}  &  \cellcolor{lightgreen}False &  \cellcolor{lightgreen}0.995 &  \cellcolor{lightgreen}0.995 &  \cellcolor{lightgreen}0.994 &  \cellcolor{lightgreen}0.994 &  \cellcolor{lightgreen}0.994 &  \cellcolor{lightgreen}0.994 &  \cellcolor{lightgreen}0.994 &  \cellcolor{lightgreen} 0.994 \\  \cline{4-12}
   &     &     &  \cellcolor{darkgreen}True &  \cellcolor{darkgreen}0.995 &  \cellcolor{darkgreen}0.89 &  \cellcolor{darkgreen}0.838 &  \cellcolor{darkgreen}0.789 &  \cellcolor{darkgreen}0.732 &  \cellcolor{darkgreen}0.678 &  \cellcolor{darkgreen}0.332 &  \cellcolor{darkgreen} 0.237 \\  \hline

    \end{tabular}
    \caption{
    Comprehensive evaluation of CW and DeepFool attack effectiveness against different defenses, showing accuracy at various L2 norm constraint values. Results demonstrate that these attacks, which find minimal perturbations, are generally less effective than bounded attacks like iFGSM and PGD.
    }
    \label{tab:l2_attacks_full}
\end{table}
\label{sec:table_linf}
\begin{table}
    \centering
    \small
    \begin{tabular}{|c|c|c|c|c|c|c|c|c|c|c|}
        \hline 
        \multicolumn{5}{|c|}{Attack parameters} & \multicolumn{6}{c|}{$l_\infty$ norm value}\\
        \hline
        Attack & Model & Defense & Through & Baseline& $\frac{2}{255}$&$\frac{4}{255}$&$\frac{6}{255}$&$\frac{8}{255}$&$\frac{10}{255}$&$\frac{12}{255}$\\
        \hline
        \multirow{14}{*}{FGSM}  &  \multirow{7}{*}{ResNet50}  &  None  &  \cellcolor{lightgray}False &  \cellcolor{lightgray}0.998 &  \cellcolor{lightgray}0.923 &  \cellcolor{lightgray}0.922 &  \cellcolor{lightgray}0.926 &  \cellcolor{lightgray}0.929 &  \cellcolor{lightgray}0.935 &  \cellcolor{lightgray} 0.937 \\  \cline{3-11}
   &     &  \multirow{2}{*}{jpeg}  &  \cellcolor{lightblue}False &  \cellcolor{lightblue}0.996 &  \cellcolor{lightblue}0.994 &  \cellcolor{lightblue}0.99 &  \cellcolor{lightblue}0.984 &  \cellcolor{lightblue}0.979 &  \cellcolor{lightblue}0.974 &  \cellcolor{lightblue} 0.971 \\  \cline{4-11}
   &     &     &  \cellcolor{darkblue}True &  \cellcolor{darkblue}0.996 &  \cellcolor{darkblue}0.908 &  \cellcolor{darkblue}0.898 &  \cellcolor{darkblue}0.893 &  \cellcolor{darkblue}0.895 &  \cellcolor{darkblue}0.893 &  \cellcolor{darkblue} 0.894 \\  \cline{3-11}
   &     &  \multirow{2}{*}{ELIC}  &  \cellcolor{lightred}False &  \cellcolor{lightred}0.983 &  \cellcolor{lightred}0.983 &  \cellcolor{lightred}0.979 &  \cellcolor{lightred}0.981 &  \cellcolor{lightred}0.98 &  \cellcolor{lightred}0.979 &  \cellcolor{lightred} 0.98 \\  \cline{4-11}
   &     &     &  \cellcolor{darkred}True &  \cellcolor{darkred}0.983 &  \cellcolor{darkred}0.921 &  \cellcolor{darkred}0.868 &  \cellcolor{darkred}0.837 &  \cellcolor{darkred}0.821 &  \cellcolor{darkred}0.809 &  \cellcolor{darkred} 0.798 \\  \cline{3-11}
   &     &  \multirow{2}{*}{HiFiC}  &  \cellcolor{lightgreen}False &  \cellcolor{lightgreen}0.975 &  \cellcolor{lightgreen}0.973 &  \cellcolor{lightgreen}0.968 &  \cellcolor{lightgreen}0.963 &  \cellcolor{lightgreen}0.958 &  \cellcolor{lightgreen}0.953 &  \cellcolor{lightgreen} 0.949 \\  \cline{4-11}
   &     &     &  \cellcolor{darkgreen}True &  \cellcolor{darkgreen}0.975 &  \cellcolor{darkgreen}0.856 &  \cellcolor{darkgreen}0.805 &  \cellcolor{darkgreen}0.786 &  \cellcolor{darkgreen}0.775 &  \cellcolor{darkgreen}0.764 &  \cellcolor{darkgreen} 0.759 \\  \cline{2-11}
   &  \multirow{7}{*}{ViT}  &  None  &  \cellcolor{lightgray}False &  \cellcolor{lightgray}0.999 &  \cellcolor{lightgray}0.911 &  \cellcolor{lightgray}0.865 &  \cellcolor{lightgray}0.843 &  \cellcolor{lightgray}0.831 &  \cellcolor{lightgray}0.828 &  \cellcolor{lightgray} 0.823 \\  \cline{3-11}
   &     &  \multirow{2}{*}{jpeg}  &  \cellcolor{lightblue}False &  \cellcolor{lightblue}0.998 &  \cellcolor{lightblue}0.978 &  \cellcolor{lightblue}0.943 &  \cellcolor{lightblue}0.909 &  \cellcolor{lightblue}0.885 &  \cellcolor{lightblue}0.872 &  \cellcolor{lightblue} 0.869 \\  \cline{4-11}
   &     &     &  \cellcolor{darkblue}True &  \cellcolor{darkblue}0.998 &  \cellcolor{darkblue}0.872 &  \cellcolor{darkblue}0.836 &  \cellcolor{darkblue}0.815 &  \cellcolor{darkblue}0.8 &  \cellcolor{darkblue}0.791 &  \cellcolor{darkblue} 0.786 \\  \cline{3-11}
   &     &  \multirow{2}{*}{ELIC}  &  \cellcolor{lightred}False &  \cellcolor{lightred}0.992 &  \cellcolor{lightred}0.989 &  \cellcolor{lightred}0.985 &  \cellcolor{lightred}0.975 &  \cellcolor{lightred}0.965 &  \cellcolor{lightred}0.952 &  \cellcolor{lightred} 0.935 \\  \cline{4-11}
   &     &     &  \cellcolor{darkred}True &  \cellcolor{darkred}0.992 &  \cellcolor{darkred}0.915 &  \cellcolor{darkred}0.821 &  \cellcolor{darkred}0.773 &  \cellcolor{darkred}0.738 &  \cellcolor{darkred}0.723 &  \cellcolor{darkred} 0.714 \\  \cline{3-11}
   &     &  \multirow{2}{*}{HiFiC}  &  \cellcolor{lightgreen}False &  \cellcolor{lightgreen}0.995 &  \cellcolor{lightgreen}0.993 &  \cellcolor{lightgreen}0.991 &  \cellcolor{lightgreen}0.984 &  \cellcolor{lightgreen}0.974 &  \cellcolor{lightgreen}0.964 &  \cellcolor{lightgreen} 0.956 \\  \cline{4-11}
   &     &     &  \cellcolor{darkgreen}True &  \cellcolor{darkgreen}0.995 &  \cellcolor{darkgreen}0.879 &  \cellcolor{darkgreen}0.824 &  \cellcolor{darkgreen}0.792 &  \cellcolor{darkgreen}0.783 &  \cellcolor{darkgreen}0.774 &  \cellcolor{darkgreen} 0.769 \\  \hline
\multirow{14}{*}{iFGSM}  &  \multirow{7}{*}{ResNet50}  &  None  &  \cellcolor{lightgray}False &  \cellcolor{lightgray}0.998 &  \cellcolor{lightgray}0.111 &  \cellcolor{lightgray}0.065 &  \cellcolor{lightgray}0.052 &  \cellcolor{lightgray}0.04 &  \cellcolor{lightgray}0.031 &  \cellcolor{lightgray} 0.024 \\  \cline{3-11}
   &     &  \multirow{2}{*}{jpeg}  &  \cellcolor{lightblue}False &  \cellcolor{lightblue}0.996 &  \cellcolor{lightblue}0.994 &  \cellcolor{lightblue}0.992 &  \cellcolor{lightblue}0.989 &  \cellcolor{lightblue}0.985 &  \cellcolor{lightblue}0.979 &  \cellcolor{lightblue} 0.971 \\  \cline{4-11}
   &     &     &  \cellcolor{darkblue}True &  \cellcolor{darkblue}0.996 &  \cellcolor{darkblue}0.423 &  \cellcolor{darkblue}0.29 &  \cellcolor{darkblue}0.237 &  \cellcolor{darkblue}0.21 &  \cellcolor{darkblue}0.185 &  \cellcolor{darkblue} 0.168 \\  \cline{3-11}
   &     &  \multirow{2}{*}{ELIC}  &  \cellcolor{lightred}False &  \cellcolor{lightred}0.983 &  \cellcolor{lightred}0.982 &  \cellcolor{lightred}0.98 &  \cellcolor{lightred}0.98 &  \cellcolor{lightred}0.981 &  \cellcolor{lightred}0.979 &  \cellcolor{lightred} 0.977 \\  \cline{4-11}
   &     &     &  \cellcolor{darkred}True &  \cellcolor{darkred}0.983 &  \cellcolor{darkred}0.785 &  \cellcolor{darkred}0.538 &  \cellcolor{darkred}0.385 &  \cellcolor{darkred}0.293 &  \cellcolor{darkred}0.221 &  \cellcolor{darkred} 0.178 \\  \cline{3-11}
   &     &  \multirow{2}{*}{HiFiC}  &  \cellcolor{lightgreen}False &  \cellcolor{lightgreen}0.975 &  \cellcolor{lightgreen}0.974 &  \cellcolor{lightgreen}0.971 &  \cellcolor{lightgreen}0.966 &  \cellcolor{lightgreen}0.962 &  \cellcolor{lightgreen}0.958 &  \cellcolor{lightgreen} 0.958 \\  \cline{4-11}
   &     &     &  \cellcolor{darkgreen}True &  \cellcolor{darkgreen}0.975 &  \cellcolor{darkgreen}0.613 &  \cellcolor{darkgreen}0.379 &  \cellcolor{darkgreen}0.259 &  \cellcolor{darkgreen}0.201 &  \cellcolor{darkgreen}0.159 &  \cellcolor{darkgreen} 0.125 \\  \cline{2-11}
   &  \multirow{7}{*}{ViT}  &  None  &  \cellcolor{lightgray}False &  \cellcolor{lightgray}0.999 &  \cellcolor{lightgray}0.158 &  \cellcolor{lightgray}0.035 &  \cellcolor{lightgray}0.013 &  \cellcolor{lightgray}0.006 &  \cellcolor{lightgray}0.004 &  \cellcolor{lightgray} 0.002 \\  \cline{3-11}
   &     &  \multirow{2}{*}{jpeg}  &  \cellcolor{lightblue}False &  \cellcolor{lightblue}0.998 &  \cellcolor{lightblue}0.975 &  \cellcolor{lightblue}0.895 &  \cellcolor{lightblue}0.746 &  \cellcolor{lightblue}0.565 &  \cellcolor{lightblue}0.395 &  \cellcolor{lightblue} 0.263 \\  \cline{4-11}
   &     &     &  \cellcolor{darkblue}True &  \cellcolor{darkblue}0.998 &  \cellcolor{darkblue}0.278 &  \cellcolor{darkblue}0.114 &  \cellcolor{darkblue}0.058 &  \cellcolor{darkblue}0.041 &  \cellcolor{darkblue}0.024 &  \cellcolor{darkblue} 0.018 \\  \cline{3-11}
   &     &  \multirow{2}{*}{ELIC}  &  \cellcolor{lightred}False &  \cellcolor{lightred}0.992 &  \cellcolor{lightred}0.986 &  \cellcolor{lightred}0.981 &  \cellcolor{lightred}0.975 &  \cellcolor{lightred}0.968 &  \cellcolor{lightred}0.959 &  \cellcolor{lightred} 0.947 \\  \cline{4-11}
   &     &     &  \cellcolor{darkred}True &  \cellcolor{darkred}0.992 &  \cellcolor{darkred}0.79 &  \cellcolor{darkred}0.5 &  \cellcolor{darkred}0.311 &  \cellcolor{darkred}0.196 &  \cellcolor{darkred}0.127 &  \cellcolor{darkred} 0.086 \\  \cline{3-11}
   &     &  \multirow{2}{*}{HiFiC}  &  \cellcolor{lightgreen}False &  \cellcolor{lightgreen}0.995 &  \cellcolor{lightgreen}0.993 &  \cellcolor{lightgreen}0.991 &  \cellcolor{lightgreen}0.986 &  \cellcolor{lightgreen}0.984 &  \cellcolor{lightgreen}0.974 &  \cellcolor{lightgreen} 0.967 \\  \cline{4-11}
   &     &     &  \cellcolor{darkgreen}True &  \cellcolor{darkgreen}0.995 &  \cellcolor{darkgreen}0.567 &  \cellcolor{darkgreen}0.262 &  \cellcolor{darkgreen}0.13 &  \cellcolor{darkgreen}0.08 &  \cellcolor{darkgreen}0.051 &  \cellcolor{darkgreen} 0.036 \\  \hline
\multirow{14}{*}{PGD}  &  \multirow{7}{*}{ResNet50}  &  None  &  \cellcolor{lightgray}False &  \cellcolor{lightgray}0.998 &  \cellcolor{lightgray}0.13 &  \cellcolor{lightgray}0.073 &  \cellcolor{lightgray}0.051 &  \cellcolor{lightgray}0.037 &  \cellcolor{lightgray}0.033 &  \cellcolor{lightgray} 0.026 \\  \cline{3-11}
   &     &  \multirow{2}{*}{jpeg}  &  \cellcolor{lightblue}False &  \cellcolor{lightblue}0.996 &  \cellcolor{lightblue}0.994 &  \cellcolor{lightblue}0.993 &  \cellcolor{lightblue}0.99 &  \cellcolor{lightblue}0.986 &  \cellcolor{lightblue}0.981 &  \cellcolor{lightblue} 0.975 \\  \cline{4-11}
   &     &     &  \cellcolor{darkblue}True &  \cellcolor{darkblue}0.996 &  \cellcolor{darkblue}0.445 &  \cellcolor{darkblue}0.307 &  \cellcolor{darkblue}0.253 &  \cellcolor{darkblue}0.218 &  \cellcolor{darkblue}0.189 &  \cellcolor{darkblue} 0.169 \\  \cline{3-11}
   &     &  \multirow{2}{*}{ELIC}  &  \cellcolor{lightred}False &  \cellcolor{lightred}0.983 &  \cellcolor{lightred}0.982 &  \cellcolor{lightred}0.983 &  \cellcolor{lightred}0.979 &  \cellcolor{lightred}0.98 &  \cellcolor{lightred}0.979 &  \cellcolor{lightred} 0.977 \\  \cline{4-11}
   &     &     &  \cellcolor{darkred}True &  \cellcolor{darkred}0.983 &  \cellcolor{darkred}0.804 &  \cellcolor{darkred}0.58 &  \cellcolor{darkred}0.425 &  \cellcolor{darkred}0.321 &  \cellcolor{darkred}0.262 &  \cellcolor{darkred} 0.215 \\  \cline{3-11}
   &     &  \multirow{2}{*}{HiFiC}  &  \cellcolor{lightgreen}False &  \cellcolor{lightgreen}0.975 &  \cellcolor{lightgreen}0.974 &  \cellcolor{lightgreen}0.969 &  \cellcolor{lightgreen}0.965 &  \cellcolor{lightgreen}0.963 &  \cellcolor{lightgreen}0.961 &  \cellcolor{lightgreen} 0.955 \\  \cline{4-11}
   &     &     &  \cellcolor{darkgreen}True &  \cellcolor{darkgreen}0.975 &  \cellcolor{darkgreen}0.635 &  \cellcolor{darkgreen}0.41 &  \cellcolor{darkgreen}0.281 &  \cellcolor{darkgreen}0.216 &  \cellcolor{darkgreen}0.171 &  \cellcolor{darkgreen} 0.133 \\  \cline{2-11}
   &  \multirow{7}{*}{ViT}  &  None  &  \cellcolor{lightgray}False &  \cellcolor{lightgray}0.999 &  \cellcolor{lightgray}0.176 &  \cellcolor{lightgray}0.042 &  \cellcolor{lightgray}0.016 &  \cellcolor{lightgray}0.008 &  \cellcolor{lightgray}0.003 &  \cellcolor{lightgray} 0.003 \\  \cline{3-11}
   &     &  \multirow{2}{*}{jpeg}  &  \cellcolor{lightblue}False &  \cellcolor{lightblue}0.998 &  \cellcolor{lightblue}0.98 &  \cellcolor{lightblue}0.916 &  \cellcolor{lightblue}0.801 &  \cellcolor{lightblue}0.646 &  \cellcolor{lightblue}0.498 &  \cellcolor{lightblue} 0.352 \\  \cline{4-11}
   &     &     &  \cellcolor{darkblue}True &  \cellcolor{darkblue}0.998 &  \cellcolor{darkblue}0.304 &  \cellcolor{darkblue}0.121 &  \cellcolor{darkblue}0.068 &  \cellcolor{darkblue}0.037 &  \cellcolor{darkblue}0.026 &  \cellcolor{darkblue} 0.017 \\  \cline{3-11}
   &     &  \multirow{2}{*}{ELIC}  &  \cellcolor{lightred}False &  \cellcolor{lightred}0.992 &  \cellcolor{lightred}0.986 &  \cellcolor{lightred}0.981 &  \cellcolor{lightred}0.977 &  \cellcolor{lightred}0.971 &  \cellcolor{lightred}0.964 &  \cellcolor{lightred} 0.958 \\  \cline{4-11}
   &     &     &  \cellcolor{darkred}True &  \cellcolor{darkred}0.992 &  \cellcolor{darkred}0.801 &  \cellcolor{darkred}0.523 &  \cellcolor{darkred}0.348 &  \cellcolor{darkred}0.23 &  \cellcolor{darkred}0.16 &  \cellcolor{darkred} 0.11 \\  \cline{3-11}
   &     &  \multirow{2}{*}{HiFiC}  &  \cellcolor{lightgreen}False &  \cellcolor{lightgreen}0.995 &  \cellcolor{lightgreen}0.993 &  \cellcolor{lightgreen}0.992 &  \cellcolor{lightgreen}0.988 &  \cellcolor{lightgreen}0.984 &  \cellcolor{lightgreen}0.978 &  \cellcolor{lightgreen} 0.974 \\  \cline{4-11}
   &     &     &  \cellcolor{darkgreen}True &  \cellcolor{darkgreen}0.995 &  \cellcolor{darkgreen}0.604 &  \cellcolor{darkgreen}0.297 &  \cellcolor{darkgreen}0.164 &  \cellcolor{darkgreen}0.1 &  \cellcolor{darkgreen}0.066 &  \cellcolor{darkgreen} 0.046 \\  \hline
    \end{tabular}
    \caption{
    Comprehensive evaluation of FGSM, iFGSM, and PGD attack effectiveness against different defenses, showing accuracy at various L$\infty$ norm constraint values. The table highlights the superior performance of learned compression methods, particularly for the ViT architecture.
    }
    \label{tab:linf_attacks_full}
\end{table}

\section{Supplementing tables on quality ablation}
\label{sec:T_ablation}

\begin{table}[H]
    \centering
    \small
    \begin{tabular}{|c|c|c|c|c|c|c|c|}
        \hline 
        \multicolumn{4}{|c|}{Attack parameters} & \multicolumn{4}{c|}{$l_2$ norm value}\\
        \hline
        Model & Quality & Through & Baseline& 2&4&8&12\\
        \hline
        \multirow{16}{*}{ResNet50}  &  \multirow{2}{*}{5.0}  &  \cellcolor{lightgray}False &  \cellcolor{lightgray}0.931 &  \cellcolor{lightgray}0.933 &  \cellcolor{lightgray}0.932 &  \cellcolor{lightgray}0.927 &  \cellcolor{lightgray} 0.929 \\  \cline{3-8}
   &     &  \cellcolor{gray}True &  \cellcolor{gray}0.931 &  \cellcolor{gray}0.384 &  \cellcolor{gray}0.224 &  \cellcolor{gray}0.145 &  \cellcolor{gray} 0.114 \\  \cline{2-8}
   &  \multirow{2}{*}{10.0}  &  \cellcolor{lightgray}False &  \cellcolor{lightgray}0.985 &  \cellcolor{lightgray}0.983 &  \cellcolor{lightgray}0.982 &  \cellcolor{lightgray}0.979 &  \cellcolor{lightgray} 0.977 \\  \cline{3-8}
   &     &  \cellcolor{gray}True &  \cellcolor{gray}0.985 &  \cellcolor{gray}0.451 &  \cellcolor{gray}0.292 &  \cellcolor{gray}0.205 &  \cellcolor{gray} 0.165 \\  \cline{2-8}
   &  \multirow{2}{*}{15.0}  &  \cellcolor{lightgray}False &  \cellcolor{lightgray}0.991 &  \cellcolor{lightgray}0.99 &  \cellcolor{lightgray}0.988 &  \cellcolor{lightgray}0.984 &  \cellcolor{lightgray} 0.979 \\  \cline{3-8}
   &     &  \cellcolor{gray}True &  \cellcolor{gray}0.991 &  \cellcolor{gray}0.444 &  \cellcolor{gray}0.303 &  \cellcolor{gray}0.218 &  \cellcolor{gray} 0.18 \\  \cline{2-8}
   &  \multirow{2}{*}{25.0}  &  \cellcolor{lightgray}False &  \cellcolor{lightgray}0.996 &  \cellcolor{lightgray}0.994 &  \cellcolor{lightgray}0.991 &  \cellcolor{lightgray}0.985 &  \cellcolor{lightgray} 0.972 \\  \cline{3-8}
   &     &  \cellcolor{gray}True &  \cellcolor{gray}0.996 &  \cellcolor{gray}0.421 &  \cellcolor{gray}0.289 &  \cellcolor{gray}0.205 &  \cellcolor{gray} 0.169 \\  \cline{2-8}
   &  \multirow{2}{*}{35.0}  &  \cellcolor{lightgray}False &  \cellcolor{lightgray}0.997 &  \cellcolor{lightgray}0.995 &  \cellcolor{lightgray}0.993 &  \cellcolor{lightgray}0.979 &  \cellcolor{lightgray} 0.955 \\  \cline{3-8}
   &     &  \cellcolor{gray}True &  \cellcolor{gray}0.997 &  \cellcolor{gray}0.398 &  \cellcolor{gray}0.281 &  \cellcolor{gray}0.195 &  \cellcolor{gray} 0.146 \\  \cline{2-8}
   &  \multirow{2}{*}{50.0}  &  \cellcolor{lightgray}False &  \cellcolor{lightgray}0.997 &  \cellcolor{lightgray}0.995 &  \cellcolor{lightgray}0.988 &  \cellcolor{lightgray}0.963 &  \cellcolor{lightgray} 0.911 \\  \cline{3-8}
   &     &  \cellcolor{gray}True &  \cellcolor{gray}0.997 &  \cellcolor{gray}0.384 &  \cellcolor{gray}0.258 &  \cellcolor{gray}0.182 &  \cellcolor{gray} 0.137 \\  \cline{2-8}
   &  \multirow{2}{*}{75.0}  &  \cellcolor{lightgray}False &  \cellcolor{lightgray}0.997 &  \cellcolor{lightgray}0.991 &  \cellcolor{lightgray}0.961 &  \cellcolor{lightgray}0.824 &  \cellcolor{lightgray} 0.622 \\  \cline{3-8}
   &     &  \cellcolor{gray}True &  \cellcolor{gray}0.997 &  \cellcolor{gray}0.34 &  \cellcolor{gray}0.235 &  \cellcolor{gray}0.137 &  \cellcolor{gray} 0.1 \\  \cline{2-8}
   &  \multirow{2}{*}{95.0}  &  \cellcolor{lightgray}False &  \cellcolor{lightgray}0.998 &  \cellcolor{lightgray}0.718 &  \cellcolor{lightgray}0.343 &  \cellcolor{lightgray}0.153 &  \cellcolor{lightgray} 0.085 \\  \cline{3-8}
   &     &  \cellcolor{gray}True &  \cellcolor{gray}0.998 &  \cellcolor{gray}0.267 &  \cellcolor{gray}0.152 &  \cellcolor{gray}0.094 &  \cellcolor{gray} 0.067 \\  \hline
\multirow{16}{*}{ViT}  &  \multirow{2}{*}{5.0}  &  \cellcolor{lightgray}False &  \cellcolor{lightgray}0.94 &  \cellcolor{lightgray}0.932 &  \cellcolor{lightgray}0.925 &  \cellcolor{lightgray}0.911 &  \cellcolor{lightgray} 0.896 \\  \cline{3-8}
   &     &  \cellcolor{gray}True &  \cellcolor{gray}0.94 &  \cellcolor{gray}0.337 &  \cellcolor{gray}0.135 &  \cellcolor{gray}0.042 &  \cellcolor{gray} 0.016 \\  \cline{2-8}
   &  \multirow{2}{*}{10.0}  &  \cellcolor{lightgray}False &  \cellcolor{lightgray}0.992 &  \cellcolor{lightgray}0.983 &  \cellcolor{lightgray}0.968 &  \cellcolor{lightgray}0.925 &  \cellcolor{lightgray} 0.854 \\  \cline{3-8}
   &     &  \cellcolor{gray}True &  \cellcolor{gray}0.992 &  \cellcolor{gray}0.358 &  \cellcolor{gray}0.145 &  \cellcolor{gray}0.048 &  \cellcolor{gray} 0.021 \\  \cline{2-8}
   &  \multirow{2}{*}{15.0}  &  \cellcolor{lightgray}False &  \cellcolor{lightgray}0.996 &  \cellcolor{lightgray}0.985 &  \cellcolor{lightgray}0.958 &  \cellcolor{lightgray}0.845 &  \cellcolor{lightgray} 0.636 \\  \cline{3-8}
   &     &  \cellcolor{gray}True &  \cellcolor{gray}0.996 &  \cellcolor{gray}0.324 &  \cellcolor{gray}0.116 &  \cellcolor{gray}0.039 &  \cellcolor{gray} 0.02 \\  \cline{2-8}
   &  \multirow{2}{*}{25.0}  &  \cellcolor{lightgray}False &  \cellcolor{lightgray}0.998 &  \cellcolor{lightgray}0.974 &  \cellcolor{lightgray}0.898 &  \cellcolor{lightgray}0.561 &  \cellcolor{lightgray} 0.264 \\  \cline{3-8}
   &     &  \cellcolor{gray}True &  \cellcolor{gray}0.998 &  \cellcolor{gray}0.272 &  \cellcolor{gray}0.111 &  \cellcolor{gray}0.037 &  \cellcolor{gray} 0.018 \\  \cline{2-8}
   &  \multirow{2}{*}{35.0}  &  \cellcolor{lightgray}False &  \cellcolor{lightgray}0.999 &  \cellcolor{lightgray}0.955 &  \cellcolor{lightgray}0.786 &  \cellcolor{lightgray}0.329 &  \cellcolor{lightgray} 0.11 \\  \cline{3-8}
   &     &  \cellcolor{gray}True &  \cellcolor{gray}0.999 &  \cellcolor{gray}0.256 &  \cellcolor{gray}0.1 &  \cellcolor{gray}0.039 &  \cellcolor{gray} 0.02 \\  \cline{2-8}
   &  \multirow{2}{*}{50.0}  &  \cellcolor{lightgray}False &  \cellcolor{lightgray}0.999 &  \cellcolor{lightgray}0.914 &  \cellcolor{lightgray}0.613 &  \cellcolor{lightgray}0.159 &  \cellcolor{lightgray} 0.033 \\  \cline{3-8}
   &     &  \cellcolor{gray}True &  \cellcolor{gray}0.999 &  \cellcolor{gray}0.247 &  \cellcolor{gray}0.091 &  \cellcolor{gray}0.031 &  \cellcolor{gray} 0.016 \\  \cline{2-8}
   &  \multirow{2}{*}{75.0}  &  \cellcolor{lightgray}False &  \cellcolor{lightgray}0.999 &  \cellcolor{lightgray}0.735 &  \cellcolor{lightgray}0.253 &  \cellcolor{lightgray}0.03 &  \cellcolor{lightgray} 0.006 \\  \cline{3-8}
   &     &  \cellcolor{gray}True &  \cellcolor{gray}0.999 &  \cellcolor{gray}0.237 &  \cellcolor{gray}0.087 &  \cellcolor{gray}0.028 &  \cellcolor{gray} 0.012 \\  \cline{2-8}
   &  \multirow{2}{*}{95.0}  &  \cellcolor{lightgray}False &  \cellcolor{lightgray}0.999 &  \cellcolor{lightgray}0.333 &  \cellcolor{lightgray}0.079 &  \cellcolor{lightgray}0.012 &  \cellcolor{lightgray} 0.004 \\  \cline{3-8}
   &     &  \cellcolor{gray}True &  \cellcolor{gray}0.999 &  \cellcolor{gray}0.256 &  \cellcolor{gray}0.075 &  \cellcolor{gray}0.018 &  \cellcolor{gray} 0.006 \\  \hline
    \end{tabular}
    \caption{
    Ablation studies comparing model accuracy for different quality levels of JPEG compression defense against iFGSM attacks with varying L2 norm constraints. These results informed the selection of optimal quality settings for the main experiments.
    }
    \label{tab:abl_jpeg}
\end{table}

\begin{table}[H]
    \centering
    \small
    \begin{tabular}{|c|c|c|c|c|c|c|c|}
        \hline 
        \multicolumn{4}{|c|}{Attack parameters} & \multicolumn{4}{c|}{$l_2$ norm value}\\
        \hline
        Model & Quality & Through & Baseline& 2&4&8&12\\
        \hline
        \multirow{6}{*}{ResNet50}  &  \multirow{2}{*}{low}  &  \cellcolor{lightgray}False &  \cellcolor{lightgray}0.976 &  \cellcolor{lightgray}0.973 &  \cellcolor{lightgray}0.967 &  \cellcolor{lightgray}0.965 &  \cellcolor{lightgray} 0.956 \\  \cline{3-8}
   &     &  \cellcolor{gray}True &  \cellcolor{gray}0.976 &  \cellcolor{gray}0.61 &  \cellcolor{gray}0.376 &  \cellcolor{gray}0.201 &  \cellcolor{gray} 0.126 \\  \cline{2-8}
   &  \multirow{2}{*}{med}  &  \cellcolor{lightgray}False &  \cellcolor{lightgray}0.995 &  \cellcolor{lightgray}0.993 &  \cellcolor{lightgray}0.991 &  \cellcolor{lightgray}0.983 &  \cellcolor{lightgray} 0.978 \\  \cline{3-8}
   &     &  \cellcolor{gray}True &  \cellcolor{gray}0.995 &  \cellcolor{gray}0.539 &  \cellcolor{gray}0.31 &  \cellcolor{gray}0.158 &  \cellcolor{gray} 0.106 \\  \cline{2-8}
   &  \multirow{2}{*}{hi}  &  \cellcolor{lightgray}False &  \cellcolor{lightgray}0.992 &  \cellcolor{lightgray}0.991 &  \cellcolor{lightgray}0.986 &  \cellcolor{lightgray}0.978 &  \cellcolor{lightgray} 0.968 \\  \cline{3-8}
   &     &  \cellcolor{gray}True &  \cellcolor{gray}0.992 &  \cellcolor{gray}0.531 &  \cellcolor{gray}0.302 &  \cellcolor{gray}0.148 &  \cellcolor{gray} 0.085 \\  \hline
\multirow{6}{*}{ViT}  &  \multirow{2}{*}{low}  &  \cellcolor{lightgray}False &  \cellcolor{lightgray}0.995 &  \cellcolor{lightgray}0.991 &  \cellcolor{lightgray}0.99 &  \cellcolor{lightgray}0.983 &  \cellcolor{lightgray} 0.967 \\  \cline{3-8}
   &     &  \cellcolor{gray}True &  \cellcolor{gray}0.995 &  \cellcolor{gray}0.575 &  \cellcolor{gray}0.255 &  \cellcolor{gray}0.083 &  \cellcolor{gray} 0.038 \\  \cline{2-8}
   &  \multirow{2}{*}{med}  &  \cellcolor{lightgray}False &  \cellcolor{lightgray}0.999 &  \cellcolor{lightgray}0.995 &  \cellcolor{lightgray}0.984 &  \cellcolor{lightgray}0.921 &  \cellcolor{lightgray} 0.822 \\  \cline{3-8}
   &     &  \cellcolor{gray}True &  \cellcolor{gray}0.999 &  \cellcolor{gray}0.509 &  \cellcolor{gray}0.212 &  \cellcolor{gray}0.058 &  \cellcolor{gray} 0.024 \\  \cline{2-8}
   &  \multirow{2}{*}{hi}  &  \cellcolor{lightgray}False &  \cellcolor{lightgray}0.998 &  \cellcolor{lightgray}0.992 &  \cellcolor{lightgray}0.971 &  \cellcolor{lightgray}0.868 &  \cellcolor{lightgray} 0.699 \\  \cline{3-8}
   &     &  \cellcolor{gray}True &  \cellcolor{gray}0.998 &  \cellcolor{gray}0.362 &  \cellcolor{gray}0.1 &  \cellcolor{gray}0.019 &  \cellcolor{gray} 0.006 \\  \hline

    \end{tabular}
    \caption{
    Ablation studies comparing model accuracy for different quality levels of HiFiC compression defense against iFGSM attacks with varying L2 norm constraints. These results informed the selection of optimal quality settings for the main experiments.
    }
    \label{tab:abl_hific}
\end{table}

\begin{table}[H]
    \centering
    \small
    \begin{tabular}{|c|c|c|c|c|c|c|c|}
        \hline 
        \multicolumn{4}{|c|}{Attack parameters} & \multicolumn{4}{c|}{$l_2$ norm value}\\
        \hline
        Model & Quality & Through & Baseline& 2&4&8&12\\
        \hline
        \multirow{7}{*}{ResNet50}  &  \multirow{2}{*}{0004}  &  \cellcolor{lightgray}False &  \cellcolor{lightgray}0.901 &  \cellcolor{lightgray}0.902 &  \cellcolor{lightgray}0.899 &  \cellcolor{lightgray}0.897 &  \cellcolor{lightgray} 0.901 \\  \cline{3-8}
   &     &  \cellcolor{gray}True &  \cellcolor{gray}0.901 &  \cellcolor{gray}0.714 &  \cellcolor{gray}0.513 &  \cellcolor{gray}0.261 &  \cellcolor{gray} 0.154 \\  \cline{2-8}
   &  \multirow{2}{*}{0008}  &  \cellcolor{lightgray}False &  \cellcolor{lightgray}0.957 &  \cellcolor{lightgray}0.959 &  \cellcolor{lightgray}0.956 &  \cellcolor{lightgray}0.953 &  \cellcolor{lightgray} 0.955 \\  \cline{3-8}
   &     &  \cellcolor{gray}True &  \cellcolor{gray}0.956 &  \cellcolor{gray}0.761 &  \cellcolor{gray}0.534 &  \cellcolor{gray}0.275 &  \cellcolor{gray} 0.17 \\  \cline{2-8}
   &  \multirow{2}{*}{0016}  &  \cellcolor{lightgray}False &  \cellcolor{lightgray}0.983 &  \cellcolor{lightgray}0.982 &  \cellcolor{lightgray}0.981 &  \cellcolor{lightgray}0.979 &  \cellcolor{lightgray} 0.979 \\  \cline{3-8}
   &     &  \cellcolor{gray}True &  \cellcolor{gray}0.984 &  \cellcolor{gray}0.785 &  \cellcolor{gray}0.542 &  \cellcolor{gray}0.289 &  \cellcolor{gray} 0.18 \\  \cline{2-8}
   &  \multirow{2}{*}{0032}  &  \cellcolor{lightgray}False &  \cellcolor{lightgray}0.992 &  \cellcolor{lightgray}0.99 &  \cellcolor{lightgray}0.99 &  \cellcolor{lightgray}0.989 &  \cellcolor{lightgray} 0.984 \\  \cline{3-8}
   &     &  \cellcolor{gray}True &  \cellcolor{gray}0.992 &  \cellcolor{gray}0.783 &  \cellcolor{gray}0.512 &  \cellcolor{gray}0.243 &  \cellcolor{gray} 0.142 \\  \cline{2-8}
   &  \multirow{2}{*}{0150}  &  \cellcolor{lightgray}False &  \cellcolor{lightgray}0.998 &  \cellcolor{lightgray}0.997 &  \cellcolor{lightgray}0.994 &  \cellcolor{lightgray}0.984 &  \cellcolor{lightgray} 0.951 \\  \cline{3-8}
   &     &  \cellcolor{gray}True &  \cellcolor{gray}0.998 &  \cellcolor{gray}0.696 &  \cellcolor{gray}0.371 &  \cellcolor{gray}0.154 &  \cellcolor{gray} 0.092 \\  \cline{2-8}
   &  \multirow{2}{*}{0450}  &  \cellcolor{lightgray}False &  \cellcolor{lightgray}0.998 &  \cellcolor{lightgray}0.996 &  \cellcolor{lightgray}0.989 &  \cellcolor{lightgray}0.928 &  \cellcolor{lightgray} 0.775 \\  \cline{3-8}
   &     &  \cellcolor{gray}True &  \cellcolor{gray}0.998 &  \cellcolor{gray}0.57 &  \cellcolor{gray}0.255 &  \cellcolor{gray}0.111 &  \cellcolor{gray} 0.073 \\  \hline
\multirow{7}{*}{ViT}  &  \multirow{2}{*}{0004}  &  \cellcolor{lightgray}False &  \cellcolor{lightgray}0.944 &  \cellcolor{lightgray}0.933 &  \cellcolor{lightgray}0.931 &  \cellcolor{lightgray}0.924 &  \cellcolor{lightgray} 0.914 \\  \cline{3-8}
   &     &  \cellcolor{gray}True &  \cellcolor{gray}0.944 &  \cellcolor{gray}0.724 &  \cellcolor{gray}0.478 &  \cellcolor{gray}0.243 &  \cellcolor{gray} 0.139 \\  \cline{2-8}
   &  \multirow{2}{*}{0008}  &  \cellcolor{lightgray}False &  \cellcolor{lightgray}0.977 &  \cellcolor{lightgray}0.972 &  \cellcolor{lightgray}0.965 &  \cellcolor{lightgray}0.955 &  \cellcolor{lightgray} 0.947 \\  \cline{3-8}
   &     &  \cellcolor{gray}True &  \cellcolor{gray}0.977 &  \cellcolor{gray}0.765 &  \cellcolor{gray}0.514 &  \cellcolor{gray}0.236 &  \cellcolor{gray} 0.116 \\  \cline{2-8}
   &  \multirow{2}{*}{0016}  &  \cellcolor{lightgray}False &  \cellcolor{lightgray}0.993 &  \cellcolor{lightgray}0.989 &  \cellcolor{lightgray}0.982 &  \cellcolor{lightgray}0.968 &  \cellcolor{lightgray} 0.948 \\  \cline{3-8}
   &     &  \cellcolor{gray}True &  \cellcolor{gray}0.993 &  \cellcolor{gray}0.79 &  \cellcolor{gray}0.494 &  \cellcolor{gray}0.2 &  \cellcolor{gray} 0.083 \\  \cline{2-8}
   &  \multirow{2}{*}{0032}  &  \cellcolor{lightgray}False &  \cellcolor{lightgray}0.997 &  \cellcolor{lightgray}0.991 &  \cellcolor{lightgray}0.985 &  \cellcolor{lightgray}0.955 &  \cellcolor{lightgray} 0.893 \\  \cline{3-8}
   &     &  \cellcolor{gray}True &  \cellcolor{gray}0.997 &  \cellcolor{gray}0.781 &  \cellcolor{gray}0.432 &  \cellcolor{gray}0.132 &  \cellcolor{gray} 0.055 \\  \cline{2-8}
   &  \multirow{2}{*}{0150}  &  \cellcolor{lightgray}False &  \cellcolor{lightgray}0.999 &  \cellcolor{lightgray}0.989 &  \cellcolor{lightgray}0.935 &  \cellcolor{lightgray}0.675 &  \cellcolor{lightgray} 0.313 \\  \cline{3-8}
   &     &  \cellcolor{gray}True &  \cellcolor{gray}0.999 &  \cellcolor{gray}0.653 &  \cellcolor{gray}0.231 &  \cellcolor{gray}0.054 &  \cellcolor{gray} 0.02 \\  \cline{2-8}
   &  \multirow{2}{*}{0450}  &  \cellcolor{lightgray}False &  \cellcolor{lightgray}0.999 &  \cellcolor{lightgray}0.951 &  \cellcolor{lightgray}0.731 &  \cellcolor{lightgray}0.212 &  \cellcolor{lightgray} 0.038 \\  \cline{3-8}
   &     &  \cellcolor{gray}True &  \cellcolor{gray}0.999 &  \cellcolor{gray}0.495 &  \cellcolor{gray}0.139 &  \cellcolor{gray}0.032 &  \cellcolor{gray} 0.01 \\  \hline

    \end{tabular}
    \caption{
    Ablation studies comparing model accuracy for different quality levels of HiFiC compression defense against iFGSM attacks with varying L2 norm constraints. These results informed the selection of optimal quality settings for the main experiments.
    }
    \label{tab:alb_elic}
\end{table}
\label{sec:table_seq}

\begin{table}
    \centering
    \small
    \begin{tabular}{|c|c|c|c|c|} 
    \hline
        N & Baseline & $\frac{4}{255}$ & $\frac{8}{255}$ & $\frac{16}{255}$\\ \hline
        1.0 & 0.998 & 0.255 & 0.11 & 0.068 \\ \hline
        2.0 & 0.998 & 0.32 & 0.134 & 0.086 \\ \hline
        3.0 & 0.997 & 0.375 & 0.155 & 0.094 \\ \hline
        4.0 & 0.997 & 0.433 & 0.189 & 0.115 \\ \hline
        5.0 & 0.998 & 0.501 & 0.231 & 0.145 \\ \hline
        6.0 & 0.997 & 0.595 & 0.301 & 0.19 \\ \hline
        7.0 & 0.998 & 0.707 & 0.41 & 0.271 \\ \hline
    \end{tabular}
    \caption{
    Detailed results for sequential defense using ELIC, showing accuracy after N compression/decompression cycles at different attack strengths. 
    }
    \label{tab:seq_elic}
\end{table}

\begin{table}
    \centering
    \small
    \begin{tabular}{|c|c|c|c|c|} \hline
         N & Baseline & $\frac{4}{255}$ & $\frac{8}{255}$ & $\frac{16}{255}$\\ \hline
        1.0 & 0.995 & 0.312 & 0.162 & 0.101 \\ \hline
        2.0 & 0.993 & 0.359 & 0.182 & 0.116 \\ \hline
        3.0 & 0.99 & 0.399 & 0.211 & 0.135 \\ \hline
        4.0 & 0.989 & 0.444 & 0.245 & 0.153 \\ \hline
        5.0 & 0.989 & 0.498 & 0.284 & 0.177 \\ \hline
        6.0 & 0.989 & 0.549 & 0.329 & 0.213 \\ \hline
        7.0 & 0.987 & 0.602 & 0.385 & 0.253 \\ \hline
    \end{tabular}
    \caption{
    Detailed results for sequential defense using HiFiC, showing accuracy after N compression/decompression cycles at different attack strengths.
    }
    \label{tab:seq_hific}
\end{table}

\begin{table}
    \centering
    \small
    \begin{tabular}{|c|c|c|c|c|} \hline
         N & Baseline & $\frac{4}{255}$ & $\frac{8}{255}$ & $\frac{16}{255}$\\ \hline
        1.0 & 0.996 & 0.29 & 0.211 & 0.169 \\ \hline
        2.0 & 0.994 & 0.439 & 0.368 & 0.303 \\ \hline
        3.0 & 0.994 & 0.826 & 0.754 & 0.684 \\ \hline
        4.0 & 0.994 & 0.929 & 0.88 & 0.833 \\ \hline
        5.0 & 0.994 & 0.953 & 0.927 & 0.894 \\ \hline
        6.0 & 0.994 & 0.968 & 0.949 & 0.928 \\ \hline
        7.0 & 0.994 & 0.976 & 0.961 & 0.946 \\ \hline
        8.0 & 0.994 & 0.979 & 0.969 & 0.956 \\ \hline
        9.0 & 0.994 & 0.982 & 0.973 & 0.963 \\ \hline
        10.0 & 0.994 & 0.984 & 0.978 & 0.968 \\ \hline
        11.0 & 0.994 & 0.989 & 0.981 & 0.972 \\ \hline
        12.0 & 0.994 & 0.989 & 0.984 & 0.974 \\ \hline
        13.0 & 0.994 & 0.991 & 0.984 & 0.978 \\ \hline
        14.0 & 0.994 & 0.988 & 0.986 & 0.98 \\ \hline
        15.0 & 0.994 & 0.992 & 0.987 & 0.98 \\ \hline
        16.0 & 0.994 & 0.991 & 0.988 & 0.983 \\ \hline
        17.0 & 0.994 & 0.99 & 0.99 & 0.984 \\ \hline
        18.0 & 0.994 & 0.99 & 0.988 & 0.984 \\ \hline
        19.0 & 0.994 & 0.993 & 0.988 & 0.985 \\ \hline
        20.0 & 0.994 & 0.991 & 0.989 & 0.985 \\ \hline
        21.0 & 0.994 & 0.992 & 0.99 & 0.987 \\ \hline
        22.0 & 0.994 & 0.991 & 0.989 & 0.987 \\ \hline
        23.0 & 0.994 & 0.992 & 0.989 & 0.986 \\ \hline
        24.0 & 0.994 & 0.993 & 0.989 & 0.988 \\ \hline
        25.0 & 0.994 & 0.992 & 0.99 & 0.988 \\ \hline
        26.0 & 0.994 & 0.993 & 0.99 & 0.987 \\ \hline
        27.0 & 0.994 & 0.994 & 0.991 & 0.988 \\ \hline
        28.0 & 0.994 & 0.992 & 0.99 & 0.988 \\ \hline
        29.0 & 0.994 & 0.993 & 0.99 & 0.985 \\ \hline
        30.0 & 0.994 & 0.994 & 0.99 & 0.987 \\ \hline
        31.0 & 0.994 & 0.992 & 0.99 & 0.987 \\ \hline
        32.0 & 0.994 & 0.993 & 0.992 & 0.989 \\ \hline
        33.0 & 0.994 & 0.994 & 0.991 & 0.988 \\ \hline
        34.0 & 0.994 & 0.993 & 0.991 & 0.989 \\ \hline
        35.0 & 0.994 & 0.994 & 0.993 & 0.989 \\ \hline
        36.0 & 0.994 & 0.994 & 0.992 & 0.99 \\ \hline
        37.0 & 0.994 & 0.994 & 0.99 & 0.986 \\ \hline
        38.0 & 0.994 & 0.993 & 0.992 & 0.987 \\ \hline
        39.0 & 0.994 & 0.994 & 0.992 & 0.988 \\ \hline
        40.0 & 0.994 & 0.993 & 0.992 & 0.989 \\ \hline
        41.0 & 0.994 & 0.993 & 0.993 & 0.99 \\ \hline
        42.0 & 0.994 & 0.993 & 0.993 & 0.988 \\ \hline
        43.0 & 0.994 & 0.993 & 0.991 & 0.991 \\ \hline
        44.0 & 0.994 & 0.994 & 0.993 & 0.99 \\ \hline
        45.0 & 0.994 & 0.993 & 0.991 & 0.99 \\ \hline
        46.0 & 0.994 & 0.994 & 0.993 & 0.991 \\ \hline
        47.0 & 0.994 & 0.994 & 0.992 & 0.99 \\ \hline
        48.0 & 0.994 & 0.994 & 0.993 & 0.99 \\ \hline
        49.0 & 0.994 & 0.994 & 0.992 & 0.989 \\ \hline
        50.0 & 0.994 & 0.994 & 0.99 & 0.99 \\ \hline
    \end{tabular}
    \caption{
    Detailed results for sequential defense using JPEG show accuracy after N compression/decompression cycles at different attack strengths. JPEG demonstrates superior scaling with iteration count, maintaining high ($>99\%$) accuracy even after 50 cycles.
    }
    \label{tab:seq_jpeg}
\end{table}
\section{Graphs}
\FloatBarrier
\label{sec_graph_allattacks}
\begin{figure*}
    \centering
    \includegraphics[width=0.55\linewidth]{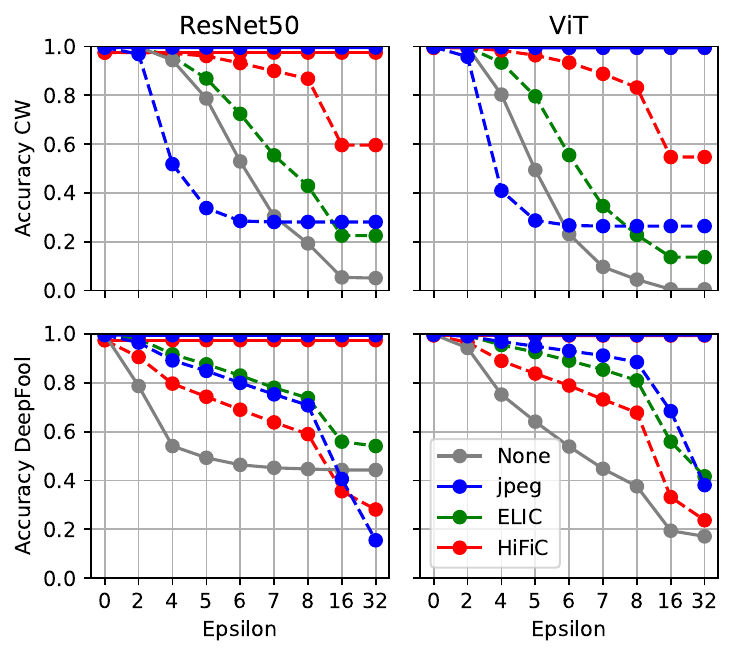}
    \caption{\Cref{fig:l_2} in a larger format}
    \label{fig:sup3}
\end{figure*}

\begin{figure*}
    \includegraphics[width=1.0\linewidth]{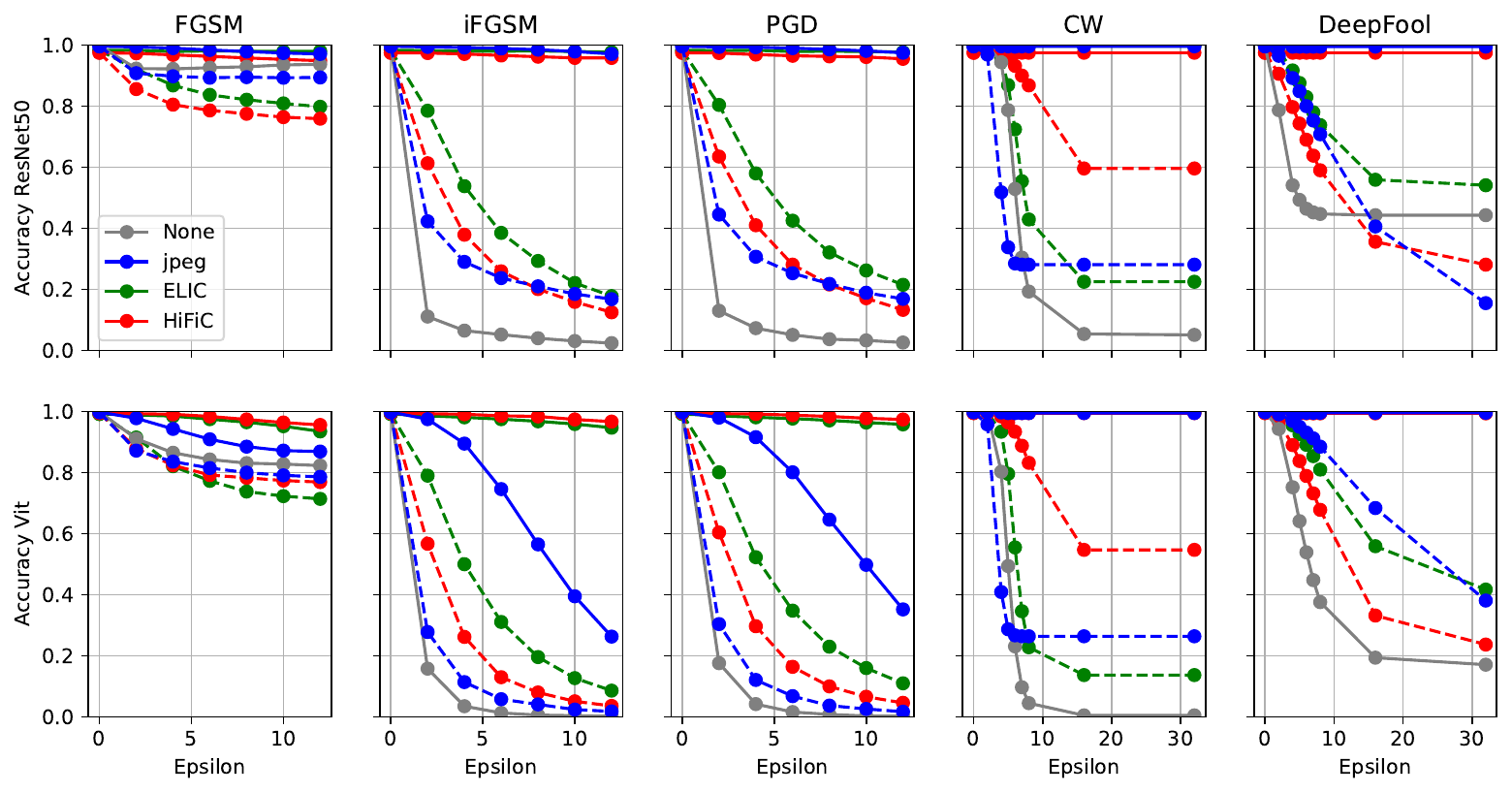}
    \caption{All attacks}
    \label{fig:sup1}
\end{figure*}

\begin{figure*}
    \centering
    \includegraphics[width=0.55\linewidth]{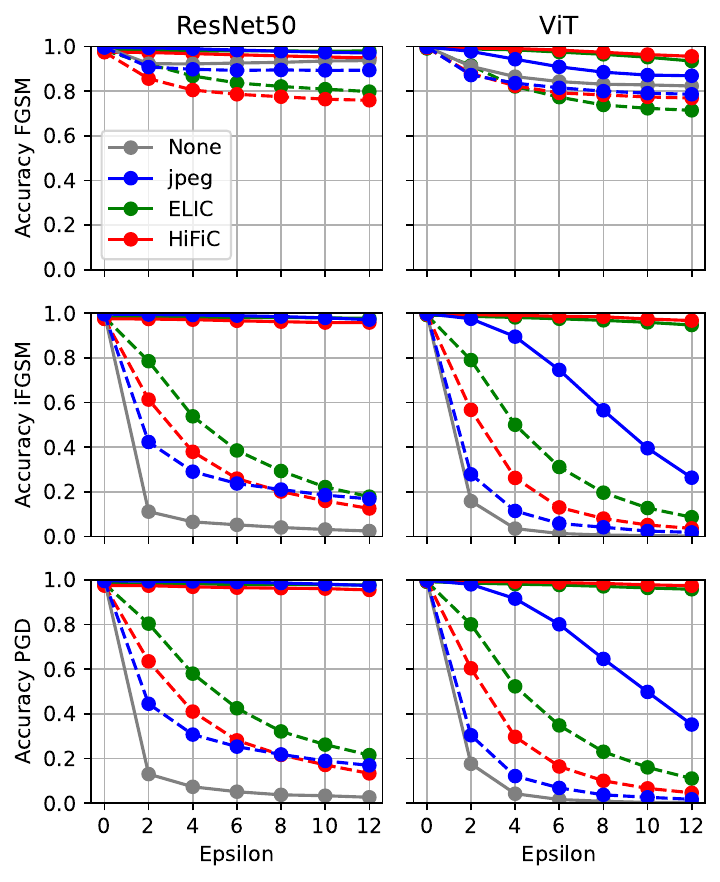}
    \caption{\Cref{fig:l_inf} in a larger format}
    \label{fig:sup2}
\end{figure*}

\begin{figure}
    \centering
    \includegraphics[width=0.5\linewidth]{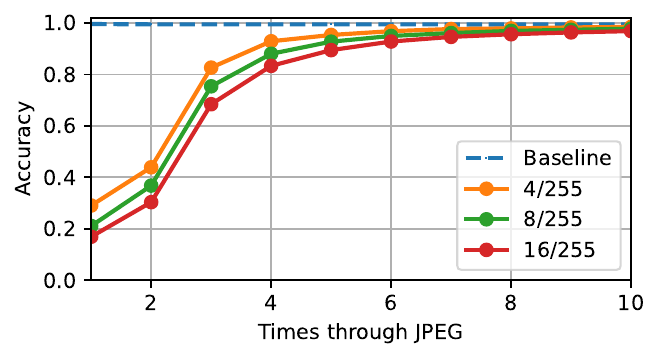}
    \caption{Accuracy of the sequential defense using multiple iterations of JPEG compression and decompression. iFGSM, ResNet50}
    \label{fig:seq-jpeg_small}
\end{figure}

\begin{figure}
    \centering
    \includegraphics[width=0.5\linewidth]{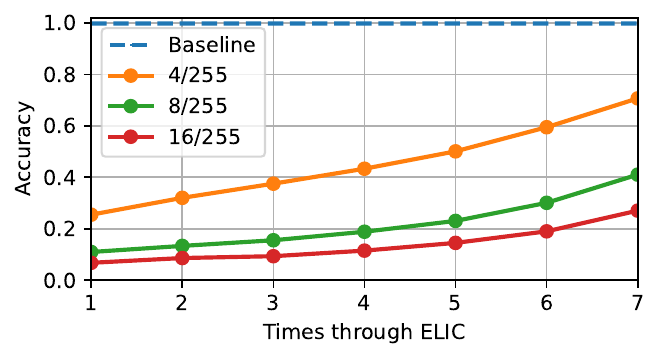}
    \caption{Accuracy of the sequential defense using multiple iterations of ELIC compression and decompression. iFGSM, ResNet50}
    \label{fig:seq-elic}
\end{figure}

\begin{figure}
    \centering
    \includegraphics[width=0.5\linewidth]{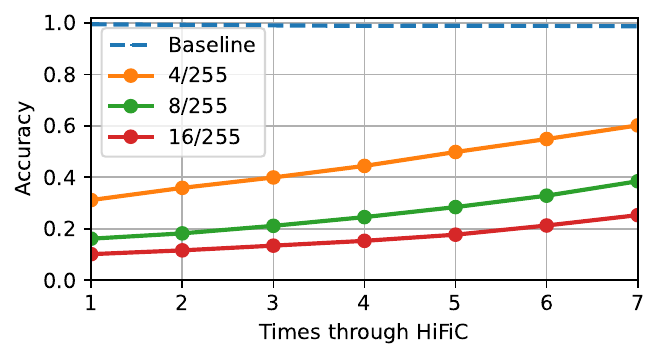}
    \caption{Accuracy of the sequential defense using multiple iterations of HiFiC compression and decompression. iFGSM, ResNet50}
    \label{fig:seq-hific}
\end{figure}

\twocolumn


\end{document}